\documentclass[journal,twocolumn,english]{IEEEtran}
\usepackage{array}
\usepackage{float}
\usepackage{pmboxdraw}
\usepackage{multirow}
\usepackage{graphicx}
\usepackage{setspace}
\usepackage{esint}
\usepackage{subcaption}
\usepackage{wrapfig}
\usepackage{amssymb}
\usepackage{amsthm}
\usepackage[T1]{fontenc}
\usepackage[utf8]{luainputenc}

\usepackage{csquotes}

\usepackage{enumitem}
\setlist[itemize]{noitemsep,topsep=0pt}
\setlist[description]{noitemsep}
\setlist[enumerate]{noitemsep}

\usepackage{tabularx}
\usepackage{booktabs}
\usepackage{multirow}
\newcolumntype{Y}{>{\raggedleft\let\newline\\\arraybackslash\hspace{0pt}}X}
\newcolumntype{Z}{>{\centering\let\newline\\\arraybackslash\hspace{0pt}}X}

\usepackage{graphicx}
\graphicspath{{images/}}
\usepackage{float}
\usepackage{caption}

\usepackage{xcolor}

\ifCLASSINFOpdf
\else
\fi

\begin{document}

\title{Towards an Efficient Semantic Segmentation Method of ID Cards for Verification Systems}


\author{Rodrigo Lara,~\IEEEmembership{Student Member,~IEEE,}
        Andres Valenzuela, Daniel Schulz, 
        Juan Tapia,~\IEEEmembership{Member,~IEEE} and Christoph Busch,~\IEEEmembership{Senior Member,~IEEE}
        \\
**\textbf{The following paper is a pre-print. The publication is currently under review for IEEE.**}        
      
\thanks{Corresponding author: Juan Tapia. Christoph Busch and Juan Tapia are with da/sec-Biometrics and Internet Security Research Group, Hochschule Darmstadt, Germany. email: christoph.busch@h-da.de and juan.tapia-farias@h-da.de. Rodrigo Lara, Andres Valenzuela, and Daniel Schulz are with TOC Biometrics, R\&D Center SR 226. emails:  rodrigo.lara, andres.valenzuela and daniel.schulz @tocbiometrics.com, Chile.}}

\maketitle

\begin{abstract}
Removing the background in ID Card images is a real challenge for remote verification systems because many of the re-digitalised images present cluttered backgrounds, poor illumination conditions, distortion and occlusions. The background in ID Card images confuses the classifiers and the text extraction.
Due to the lack of available images for research, this field represents an open problem in computer vision today.
This work proposes a method for removing the background using semantic segmentation of ID Cards. In the end, images captured in the wild from real operation, using a manually labelled dataset consisting of 45,007 images, with five types of ID Cards from three countries (Chile, Argentina and Mexico), including typical presentation attack scenarios, were used. This method can help to improve the following stages in a regular identity verification or document tampering detection system. Two Deep Learning approaches were explored, based on MobileUNet and DenseNet10. The best results were obtained using MobileUNet, with 6.5 million parameters. A Chilean ID Card's mean Intersection Over Union (IoU) was 0.9926 on a private test dataset of 4,988 images. The best results for the fused multi-country dataset of ID Card images from Chile, Argentina and Mexico reached an IoU of 0.9911. The proposed methods are lightweight enough to be used in real-time operation on mobile devices. 
\end{abstract}


\maketitle

\IEEEdisplaynontitleabstractindextext

\IEEEpeerreviewmaketitle

\section{Introduction}

\IEEEPARstart{N}{owadays}, the interest in remote digital identity verification or identification has grown drastically, boosted by the current COVID-19 pandemic. Also, with the broad use of smartphones worldwide to access different services, for example, banking, e-commerce, fintech, etc. It is critical to have a robust remote automatic verification system. One method for person verification is using identity cards (ID Card), which provide basic information about the card holder. For example, full name, date and place of birth, nationality, some identification number, signature, etc. If the ID Card contains a frontal face photograph, it can be used as the reference for a remote verification method, comparing this photo with a new one provided by the user, for example, a self portrait photograph (selfie). Therefore, ID Card, passport and driver licence images, are analysed using computer vision and Optical Character Recognition (OCR) techniques for obtaining automatically and remotely the reference information about the card holder. However, all these images are captured in unconstrained scenarios, for example, a user only need to have an smartphone camera and internet access to activate a bank account, so the captured images can present a lot of variations such as different background, illumination, geometrical deformation, focus, specular reflection, etc. The background is the external area that surrounds the ID card. 

Using ID Card analysis systems improves data input processes. The goals for this kind of systems are: Perform verification, segmentation and data extraction from documents, prevent identity fraud detecting forgeries, or classifying the document as real or fake, among others \cite{Arlazarov_2019}.

\begin{figure*}[]
\centering
\includegraphics[width=6.0in]{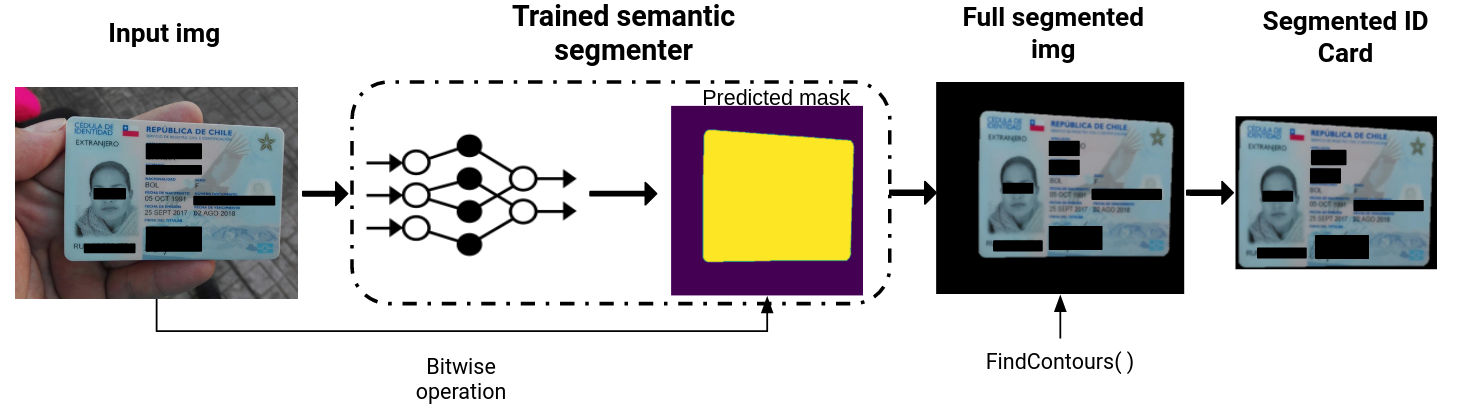}
\caption{Proposed segmentation framework for ID Cards from Chile, Argentina and Mexico.}
\label{diagrama_flujo}
\end{figure*}

In order to progress with research and development of ID Card digitisation and analysis systems, datasets of significant size are need that contain a representative collection of ID Card images. However, access to this kind of data is difficult, because of the confidential personal information that these documents contain, leading to some issues, for example, personal data leakage risk and their consequences; few people will share their personal data, knowing the risks implied; high cost for collecting ID Cards; and scarce availability of public datasets, even in some countries it is illegal to collect this kind of data. For these reasons, currently there are no publicly available datasets containing ID Cards, forcing each research team to create and maintain their own datasets with their own resources \cite{Arlazarov_2019}.

Traditionally, remote verification systems have the following stages: Image Acquisition, the card holder presents the ID Card to a smartphone in order to capture a digital photo of the entire card. Also, in this stage, a selfie of the user is captured. Afterwards, the segmentation stage removes the background from the ID Card photo. Later, in the verification stage where the selfie is compared to the ID card photo, and finally, a decision is taken or query to an external database is performed, to verify the user's identity in case the claimed identity was not successfully verified.

For the segmentation stage, the ID Card portion of the image must be isolated from the background or occlusions, in this way, the image can be analysed in further processing stages, as can be seen in Figure \ref{diagrama_flujo}. This stage must be performed because, for example, the images submitted from clients in a banking App are far from being ideal, presenting occlusions, rotations, uneven illumination, etc, or the document layout is different for each country, or even worse, each country can have many different types of ID Cards, with radically different layouts. Where in one kind, the photo of the card owner can be located in the right side (A non ICAO complaint ID Card), where in another kind the photo is in the left side. All these complexities and artefacts can confuse the system and crop the images in an inappropriate condition, hindering the following processing stages, like identity verification or reading information using OCR. 

In order to improve the segmentation stage, we propose an approach based on computer vision, specifically a semantic segmentation method, aimed to partitioning a digital image into some non-intersecting regions such that each region is homogeneous and the union of no two adjacent regions is homogeneous \cite{PAL19931277}. We use a semantic segmentation approach because only one ID Card is present in each image, then each individual pixel is assigned to a class label \cite{FCNN}, instead of other type of segmentation approach, i.e. instance segmentation. The main contributions of this work can be summarised as follows:

\begin{enumerate}
    \item This work evaluates five different ID Card types using semantic segmentation, with a dataset of 45,007 images representing ID Cards from three different countries, including real images (Digital) and three presentation attacks (Composite, Printed and Screen). This provides us great variability for training and evaluation of our models.
    \item The dataset used is sequestered due to privacy concerns. However, it will be available to other researchers for evaluating their own models by request.
    \item Both methods presented in this work, are fast enough to be used in real-time operation.
    \item This work presents results for real-life operation scenarios, with authentic images, captured using several smartphones in the wild, from retail stores, banks, etc. Therefore, it can be used as a guide for future research efforts on this topic.

\end{enumerate}

The rest of the paper is organised as follows: in section \ref{sec:related_work} related work is presented, in section \ref{sec:proposed_method} the proposed method is shown, in section \ref{sec:experimental} the experimental setup and datasets are shown, in section \ref{sec:Experiment} the experiments performed and the results obtained are detailed, finally, in section \ref{sec:conclusion} the conclusions are presented.

\section{Related work}
\label{sec:related_work}
Deep Learning and Convolutional Neural Networks (CNN) have been used successfully in the last few years, outperforming the traditional handcrafted features in many computer vision tasks, for example, image classification \cite{SimonyanZ14a, krizhevsky2012imagenet}, object detection \cite{Girshick_2015_ICCV, Ren2017, Redmon_2016_CVPR}, face recognition \cite{Taigman_2014_CVPR, Schroff_2015_CVPR, Deng_2019_CVPR}. 
Also, Deep Learning have been used for image segmentation tasks, for example, instance segmentation \cite{He_2017_ICCV}, panoptic segmentation \cite{kirillov2019panoptic}, and semantic segmentation, which have been used in many real-world applications, such as self-driving vehicles \cite{li2018real, tseng2018combination, treml2016speeding, de2017semantic, yang2018denseaspp}, pedestrian detection \cite{flohr2013pedcut, brazil2017illuminating, ullah2018pednet}, scene understanding \cite{zhou2019semantic}, defect detection \cite{tao2018automatic, crack_seg}, etc.

In the general semantic segmentation literature, we can cite the work of Long et al. \cite{FCNN}, where a classification network is transformed into a Fully Convolutional Network, taking an input of arbitrary size and producing the correspondingly sized output. This architecture recovers the spatial information from the downsampling layers adding upsampling layers to the network.
Performing efficient inference and learning, this method achieves state-of-the-art results in PASCAL VOC, NYUDv2, and SIFT Flow, with an inference time less than 1/5 second for a typical image.

In the work of Yu et al. \cite{yu2021bisenet}, a method for real-time semantic segmentation is proposed, called Bilateral Segmentation Network (BiSeNet V2), with two branches, a detail branch with wide channels and shallow layers for low-level details, and a lightweight semantics branch, with narrow channels and deep layers for high-level semantic context, obtaining state-of-the-art results on the Cityscapes dataset.

In \cite{paperUNET}, a method relying on heavy use of data augmentation is proposed, training a network end-to-end from few images, with an architecture consisting of two parts: a contracting path to capture context, and a symmetric expanding path, enabling precise localisation. This method obtained the best results on the ISBI challenge for segmentation of neuronal structures in electron microscopic stacks. 

The work of Chen et al \cite{paperDEEPLAB}, proposed a method combining the advantages of a spatial pyramid pooling module, for encoding multi-scale contextual information, and an encode-decoder structure, for capturing sharper object boundaries. This method, called DeepLabv3+, extends DeepLabv3 adding a simple decoder module to recover the object boundaries. Also, the Xception model was adapted for the semantic segmentation task, applying the depthwise separable convolution to the Atrous Spatial Pyramid Pooling and decoder modules, obtaining state-of-the-art results on PASCAL VOC 2012 and Cityscapes datasets.

In \cite{zhang2018context}, the impact of global contextual information in semantic segmentation is analyzed, introducing the Context Encoding Module and a Semantinc Encoding Loss, aimed to capture semantic context of scenes and to highlight class-dependent feature maps. Also, a new semantic segmentation framework is proposed, named Context Encoding Network (EncNet), including a Context Encoding Module into a pre-trained ResNet \cite{he2016deep}, using a dilation strategy, obtaining state-of-the-art results on PASCAL VOC 2012 and PASCAL in Context.

In \cite{zhao2017pyramid}, a Pyramid Scene Parsing Network (PSPNet) is proposed, aimed to a complete understanding of a scene, predicting the label, location and shape for each element, surpassing difficult scene context features in an traditional FCN framework. It incorporates global context information by different region-based context aggregation using a global pyramid pooling module, in this way, combining local and global clues make the prediction more reliable. Also, an effective optimization strategy for ResNet \cite{he2016deep} was developed, based on deeply supervised loss. The method obtained state-of-the-art results for scene parsing and semantic segmentation tasks.

The work of Valenzuela et al. \cite{valenzuela2020towards} proposed a method for semantic segmentation of NIR eye images, where a lightweight CNN architecture named DenseNet10, with only three blocks and ten layers is developed. The trade-off amongst grown rate (k), IoU, and the number of layers was carefully explored. The main goal was to achieve an efficient network with fewer parameters than traditional architectures in order to be used for mobile device applications.

Regarding segmentation and localisation of content elements for ID Cards, extensive literature review showed that research on this topic is limited, mostly because the low availability of real data, for the reasons mentioned in the previous section, so most of the methods uses a few examples or synthetic data. Specifically, few methods use segmentation techniques, most of them rely on estimating the quadrilateral that defines the borders of an ID Card, but this has some drawbacks, like the occlusion handling, for example, a finger that occludes a part of the ID Card. For these reasons, semantic segmentation allows a more precise localisation of all the pixels belonging to the ID Card, handling occlusions in a proper way. Also, some of the reported methods use the localisation stage as part of a pipeline for verification or type recognition (i.e. Cards from different countries), then, no comparable metrics were reported for the segmentation or localisation stage, for example, IoU.

The work of Casteblanco et al. \cite{castelblanco2020machine} studies some machine learning techniques for document verification, on a small dataset consisting of 101 Colombian ID Cards. In the Document Acquisition stage, they performed a semantic segmentation of the ID Cards using UNet, and for verification, traditional Computer Vision techniques were used for feature extraction and classification, like Histogram of Oriented Gradients, Support Vector Machine and Random Forest. In the segmentation stage, they reported 98.49\% accuracy in training, 98.41\% accuracy in test, and IoU of 0.98 in test. The main drawback of this method is the small dataset used, without enough variability, compared to a real life operational scenario.

In \cite{das2020fast}, a method based on UNet was proposed to detect document edges and text regions in Brazilian ID Card images, with a Fully Octave Convolutional Neural Network, which replaces the Convolutional Layers by Octave Convolutional Layers, reducing the redundancy of feature maps and obtaining a lighter model. In the datasets developed, the first one is named CDPhotoDataset with 20,000 images, obtaining an IoU of 0.9916; the second one is named DTDDataset, with 800 real Brazilian documents and after data augmentation a total of 10,000 images, obtaining an IoU of 0.9599.

The work of Tropin et al. \cite{tropin2021approach} proposed a combination of contour and region-
based approaches, ranking the competing contour location according to the contrast between the areas inside and outside the border. This method obtains state-of-the-art performance on the open MIDV-500 dataset, with a value of 0.97 using a variation of the Jaccard index. It is important to point out that MIDV-500 dataset presents small variability, which means that many images are from the same ID Card.

In \cite{chiron2021fast}, a method for detection, classification and alignment of identity documents is proposed, using a modular approach based on a fully multi-stage deep learning, allowing to accurately locate and classify the document. For rough detection of any kind of ID Card, a customised EfficientDet is used, followed by a classification stage based on MNASNet-A1. Finally, for fine alignment, SuperPoint and SuperGlue based approaches are used. On the MIDV500 dataset, an IoU of 98.28 was obtained, while in the private industrial dataset, a more challenging and representative of a real life application, with 14k sample images uniformly distributed over 67 different classes, an IoU of 90.43 was obtained.

The work of Awal et al. \cite{awal2017complex} proposed a method for simultaneous location and class recognition, where the classes are defined by the document nature, emission country, version, and the visible side. First, a coarse keypoint finding algorithm associates the document image to a reference model, and then, fine-grained analysis is applied for document localisation and extraction. The experiments were performed on three private datasets. No results for localisation or segmentation were reported.

In \cite{attivissimo2019automatic}, a pipeline for localisation, classification and text recognition is presented, using synthetically generated data for the main Italian identity documents, for both training and testing. For detection, the vertices of the documents are adjusted iteratively, with pixels sampled in an outer region. The main drawback of this approach for localisation, is the very heterogeneous nature of the backgrounds presents in real life operation scenarios. The vertices detection was reported as a localisation metric, obtaining an accuracy of 68.57\%.  

In \cite{tropin2021advanced}, a method based on an advanced Hough transform is proposed for detection of the quadrilateral that forms the boundaries of a document, taking into account the geometric invariants of the central projection model and combining edge and color features. This method is intended for real-time use on smartphones, without knowing any a priori information about the document content or structure. On the MIDV500 dataset, an IoU of 0.9830 with all 4 vertices within the frame was obtained.

In \cite{chiron2021id}, a method for spotting and locating identity documents in the wild is presented, using a priori information along with a list of predefined models, using specific ID document features. For solving the problem, the approach tests different crop hypotheses, competing between them, to select at least one candidate that correctly crops the document, using a custom visual similarity metric. The methods were evaluated on two datasets: MIDV500 and a private industrial dataset, with 1,587 images distributed over 79 classes belonging to 14 countries. The localisation accuracy, using a threshold for IoU of 0.9, was 92.8\% for the private dataset and 97\% for MIDV500. 

In \cite{skoryukina2019fast}, a method for simultaneous location and document type recognition is performed on ID document images. There are two considered cases, video in mobile devices, photos and scanned images on a server. For this purpose, feature points, descriptors, straight lines and quadrangles are extracted from the image. Localisation results were obtained in the MIDV500 dataset and a private dataset, yielding accuracies of 70\% and 59\%, respectively.

Finally, in a related work, Gonzalez et al. \cite{gonzalez2020hybrid} proposed a method for tampering detection on chipless ID Cards, where a two-stages CNN is developed, using BasicNet with Discrete Fourier Transform, to determine if an ID Card image provided remotely by the user is real, or tampered in the digital (composite) or non-digital domain (high-quality printed or digitally displayed on a screen). 

\section{Proposed method}
\label{sec:proposed_method}

The goal in this work is to generate a lightweight semantic segmentation method for different types of ID Cards, in order to be used on mobile devices. In this way, it is possible to obtain a more robust image for the following stages in a tampering detection, with only the pixels belonging to the ID Card being activated, without any background or occlusions. To reach this goal, we evaluated three schemes, first, a sliding window Histogram of Oriented Gradient based detector using Support Vector Machine (SVM) as classifier \cite{dalal2005histograms}; and two different CNN architectures, firstly, a MobileUNet network, and secondly, a much lighter network, that has showed good performance in NIR eye segmentation tasks, based on DenseNet10 \cite{valenzuela2020towards}.

\subsection{HOG/SVM}

The first method, serving as a baseline, is based on the work of Dalal and Triggs \cite{dalal2005histograms}, employing a sliding window approach to detect the relevant regions where the object of interest is located. For each window, a feature descriptor is calculated, called Histogram of Oriented Gradients (HOG), counting occurrences of gradient orientation in a certain portion of an image. Then, each window is classified by SVM \cite{cortes1995support}, if the window contains or does not contain the object of interest, in our case, an ID Card.

\subsection{MobileUNet}

\begin{figure}[H]
\centering
\includegraphics[scale=0.24]{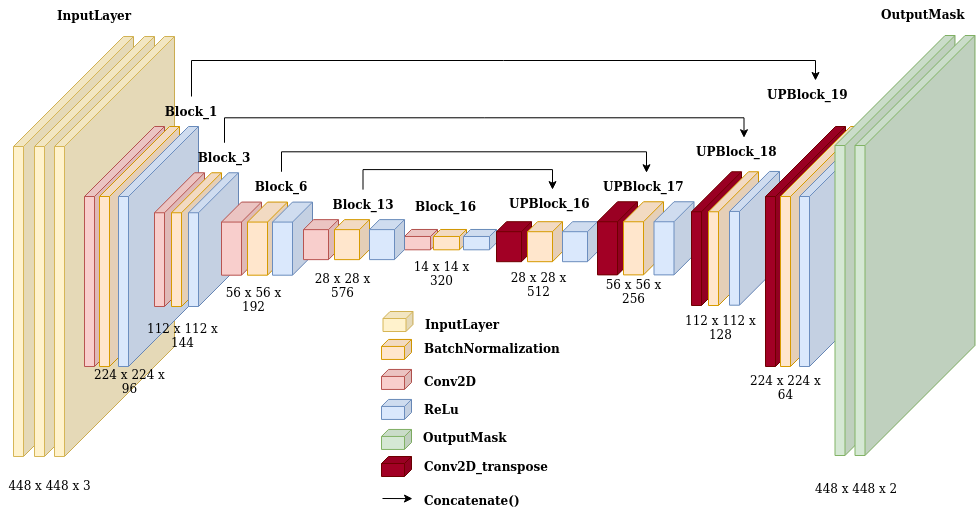}
\caption{MobileUNet CNN Network.}
\label{fig_UNET}
\end{figure}

The second method, is based on an UNet architecture, which is a CNN developed for medical images segmentation. It is composed of a contracting path (Figure \ref{fig_UNET}, left) and an expansive path (Figure \ref{fig_UNET}, right). The contracting path acts as a feature extractor, similar to VGG \cite{simonyan2014very}, consisting in the repeated application of two $3\times3$ unpadded convolutions, followed by a rectified linear unit (ReLU) and a $2\times2$ max pooling operation with stride 2 for downsampling, doubling the number of feature channels at each step. In each step of the expansive path, feature maps are upsampled, and then a $2\times2$ convolution is performed, obtaining half number of feature channels, then is concatenated with the corresponding cropped feature map from the contracting path, and two $3\times3$ convolutions followed by a ReLU. The final layer is a $1\times1$ convolution for mapping the feature vector to the desired number of classes, obtaining 23 convolutional layers for the whole network.

In our case, we wanted the network to learn robust features and reduce the number of trainable parameters. In order to do that, a MobileNetV2 architecture \cite{sandler2018mobilenetv2}, pre-trained on the ImageNet dataset \cite{russakovsky2015imagenet} is employed as the contracting path (Encoder), using their intermediate output. The UNet architecture was used as upsampling path for recovering the features of the images. The number of parameters for this method is about 6.5 millions.

\begin{figure}[H]
\centering
\includegraphics[width=3.2in]{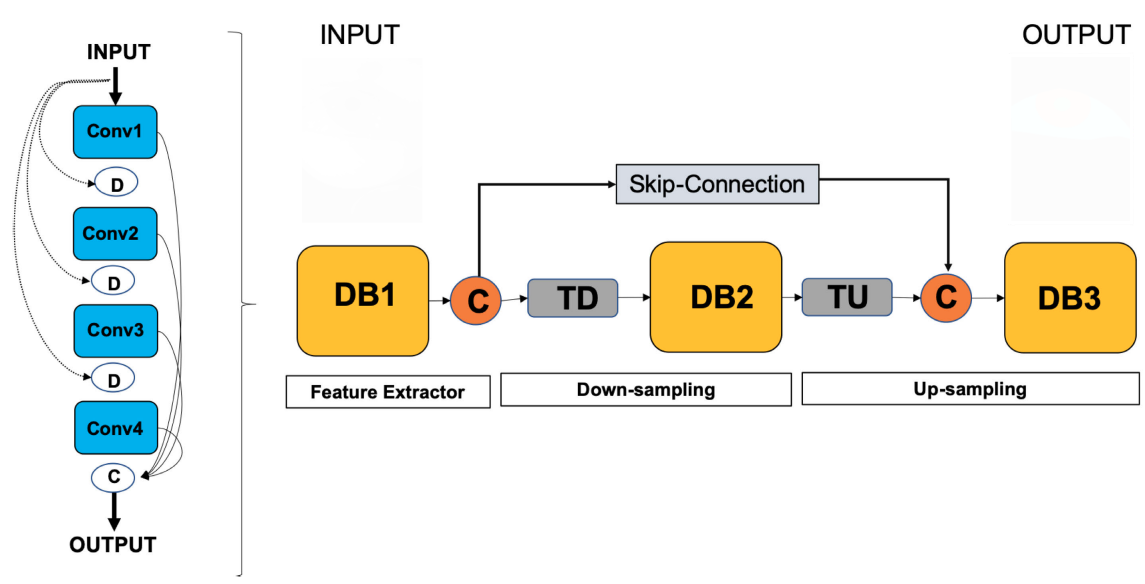}
\caption{DenseNet10 CNN network \cite{valenzuela2020towards}.}
\label{fig_densenet}
\end{figure}

\subsection{DenseNet10}

In order to improve the results obtained and reduce the number of parameters, a novel implementation of a DenseNet56 with only ten layers was proposed as follows.
This method is based on DenseNet10 \cite{valenzuela2020towards}, having only three blocks and ten layers, obtaining a lightweight and efficient network, to be used for mobile device applications. This is accomplished by having a feature extractor with two paths (Downsampling and one Upsampling), where the downsampling path has 1 Transition Down (TD) and the upsampling path has 1 Transition Up (TU), instead of the 4 Transitions (2TD+2TU) used in the traditional approach. See Figure \ref{fig_densenet}. The architecture obtained is lightweight compared to the MobileUNet approach, with only 210,732 parameters. The grown rate $K$ was set-up to $5$.

\section{Experimental setup}
\label{sec:experimental}
In this section, the performed experiments and the datasets used are explained in detail, for training and evaluation of the developed models. Also, the implementation for each network is reviewed, for each experiment. All experiments were performed on an Intel i7 6700K processor, with 64GB of RAM and a NVIDIA GTX 1080 GPU, with 8GB VRAM.
    
\subsection{Dataset}

The dataset used in the experiments contains 45,007 images with ID Cards from Chile, Argentina and Mexico, where in the two first countries, they are distributed in two different types of card per country (both in operation), according to the year of manufacturing. All the ID Card have the same size but different colours. The Chilean ID Card is build on plastic, being ICAO compliant. Conversely, Argentina and Mexican ID Card are build on hard paper and are not ICAO compliant. 

\begin{figure*}[t]
\centering
    \begin{minipage}[t]{.23\textwidth}
        \centering
        \includegraphics[width=\textwidth]{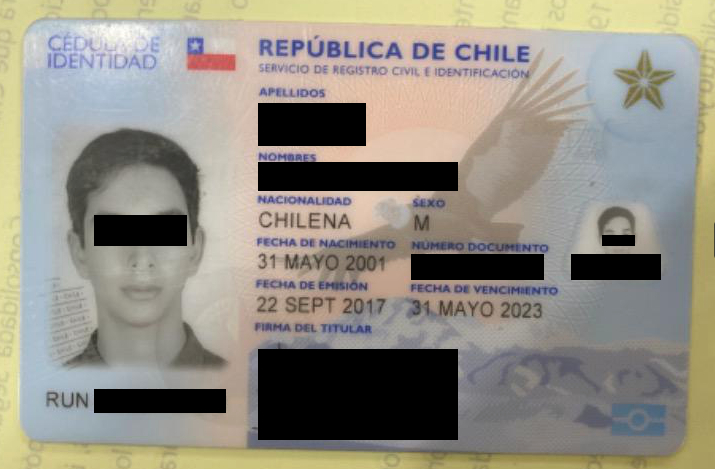}
        \subcaption{Chilean ID Card Digital.}\label{fig:1}
    \end{minipage}
    \hfill
    \begin{minipage}[t]{.245\textwidth}
        \centering
        \includegraphics[width=\textwidth]{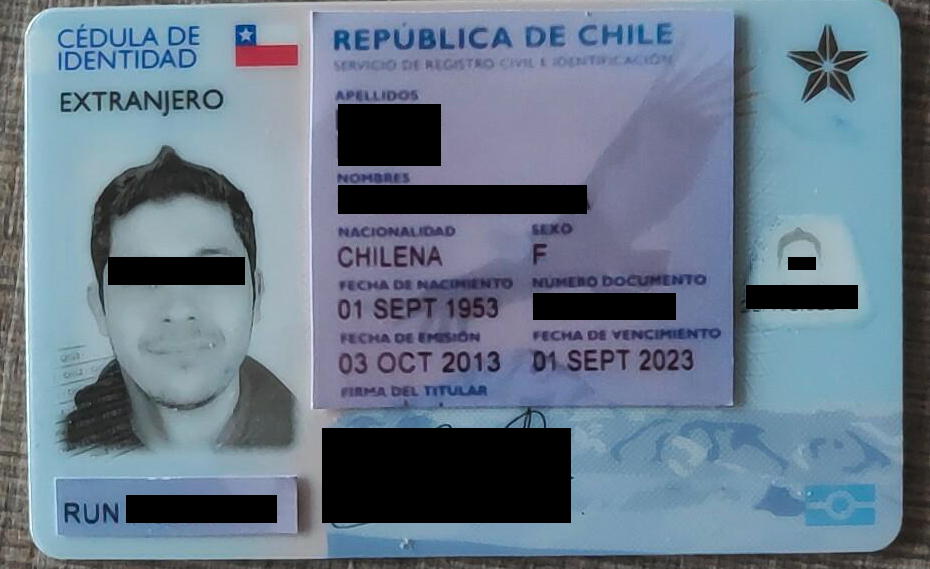}
        \subcaption{Chilean ID Card Composite.}\label{fig:2}
    \end{minipage}
    \hfill
    \begin{minipage}[t]{.25\textwidth}
        \centering
        \includegraphics[width=\textwidth]{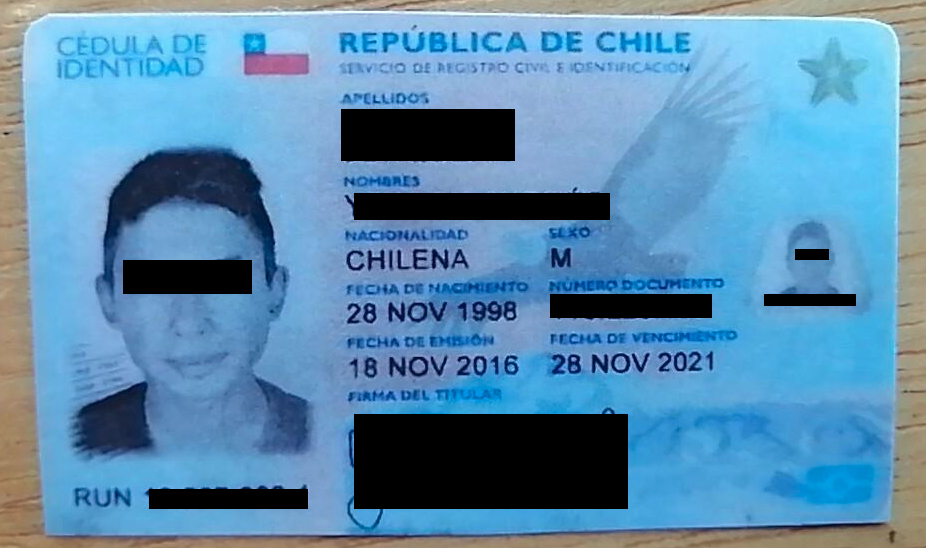}
        \subcaption{Chilean ID Card Printed.}\label{fig:3}
    \end{minipage}
    \hfill
    \begin{minipage}[t]{.22\textwidth}
        \centering
        \includegraphics[width=\textwidth]{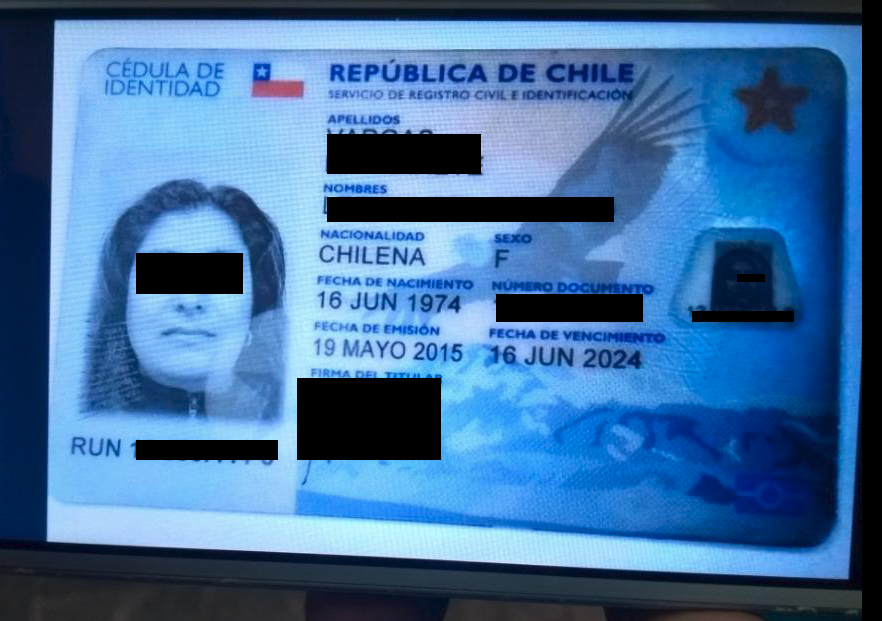}
        \subcaption{Chilean ID Card Display.}\label{fig:4}
    \end{minipage}
    
    \begin{minipage}[t]{.23\textwidth}
        \centering
        \includegraphics[width=\textwidth]{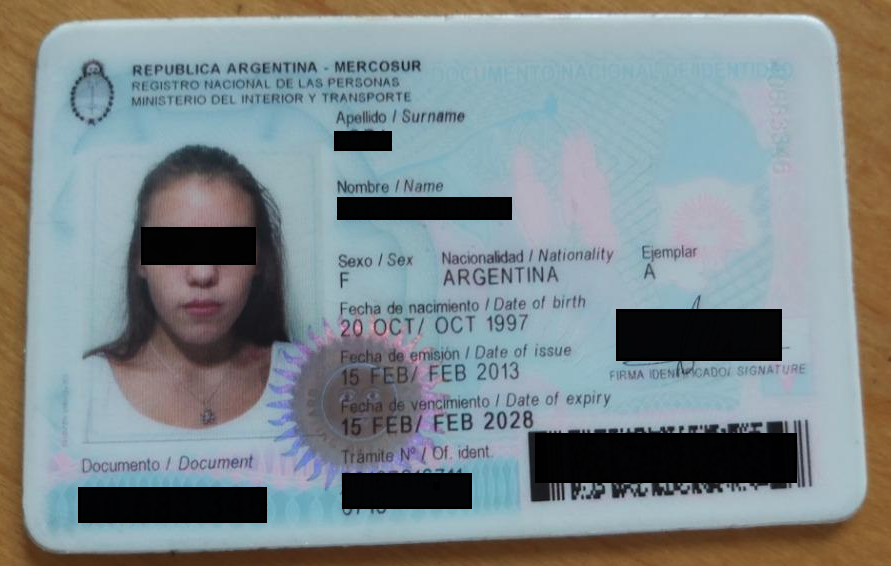}
        \subcaption{Argentinean ID Card Digital.}\label{fig:5}
    \end{minipage}
    \hfill
    \begin{minipage}[t]{.245\textwidth}
        \centering
        \includegraphics[width=\textwidth]{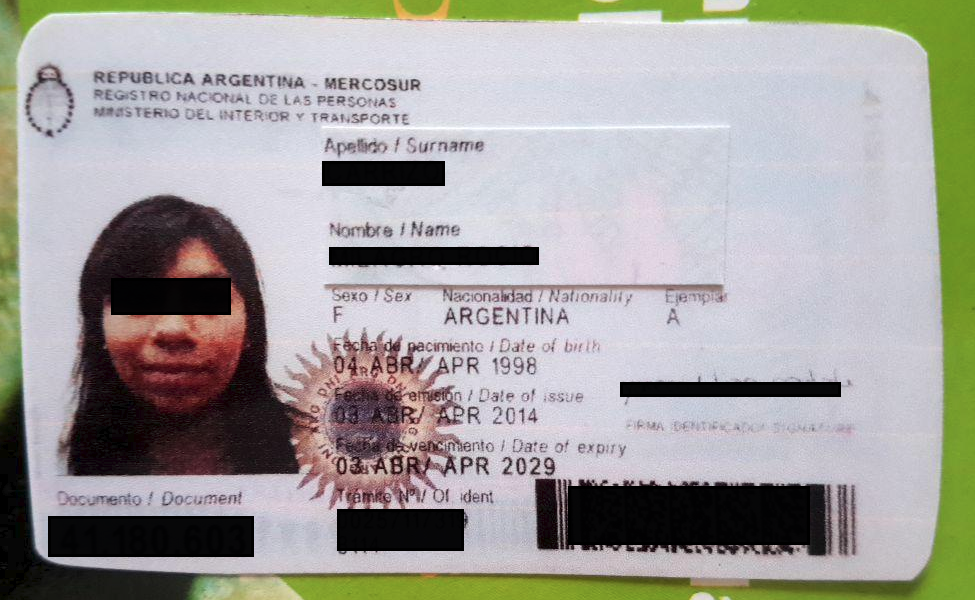}
        \subcaption{Argentinean ID Card Composite.}\label{fig:6}
    \end{minipage}
    \hfill
    \begin{minipage}[t]{.24\textwidth}
        \centering
        \includegraphics[width=\textwidth]{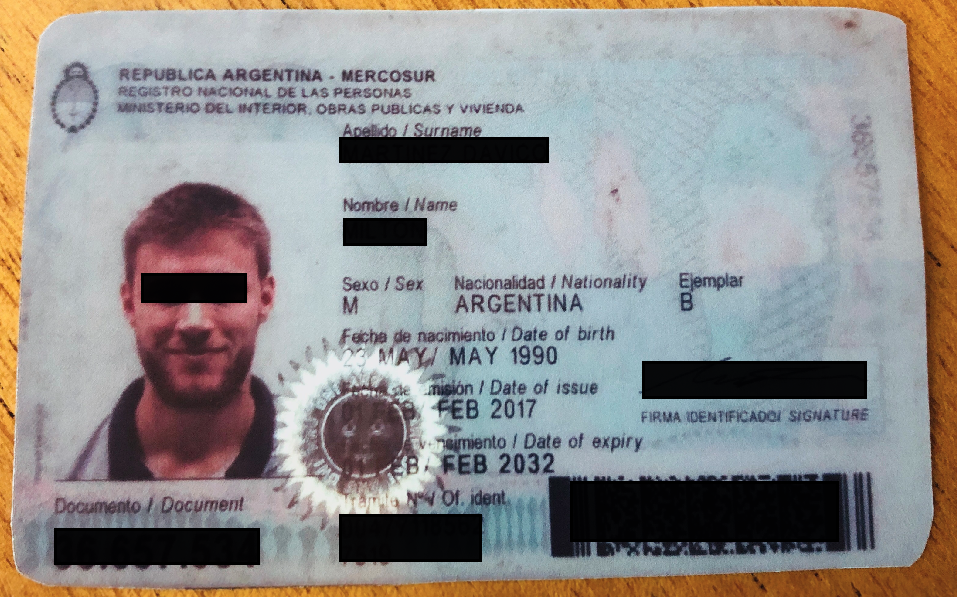}
        \subcaption{Argentinean ID Card Printed.}\label{fig:7}
    \end{minipage}
    \hfill
    \begin{minipage}[t]{.25\textwidth}
        \centering
        \includegraphics[width=\textwidth]{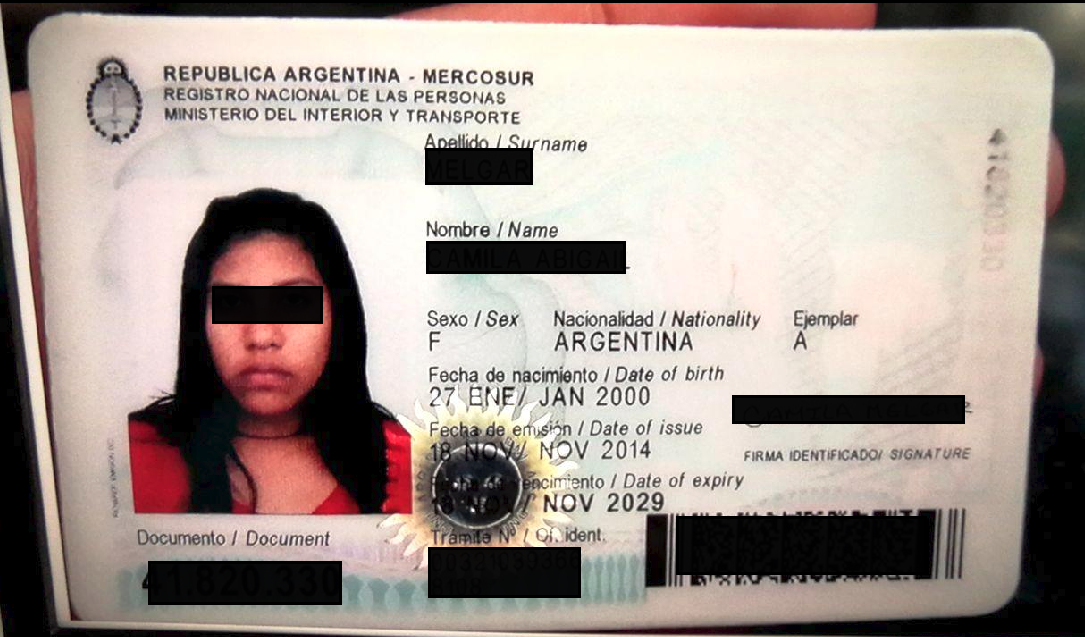}
        \subcaption{Argentinean ID Card Display.}\label{fig:8}
    \end{minipage}
    \begin{minipage}[t]{.23\textwidth}
        \centering
        \includegraphics[width=\textwidth]{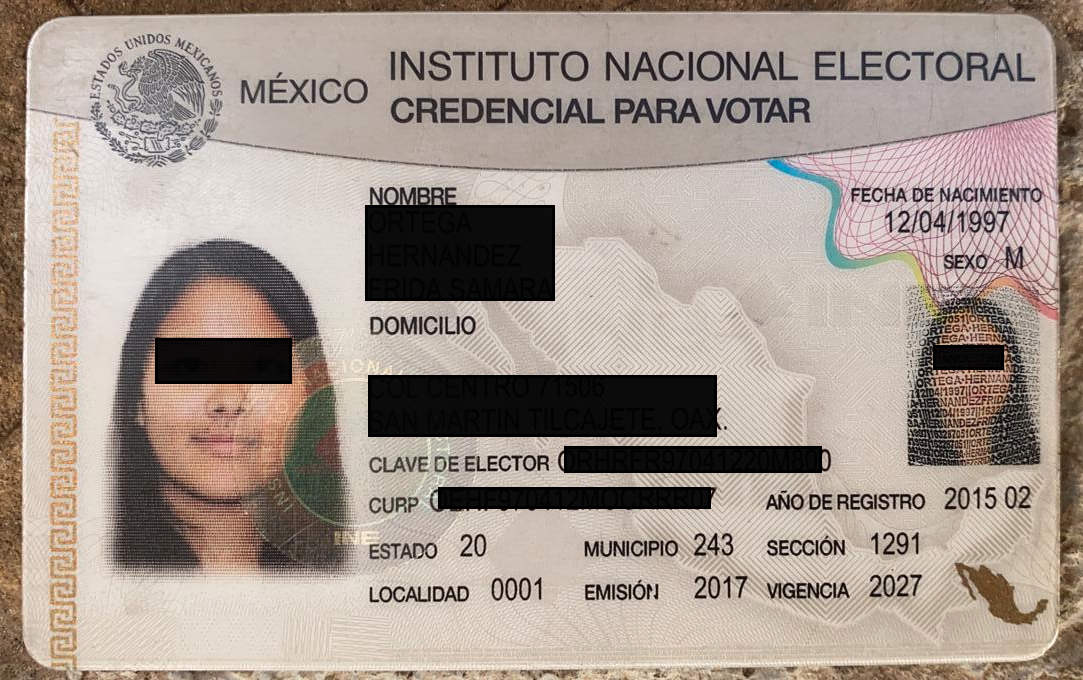}
        \subcaption{Mexican ID Card Digital.}\label{fig:9}
    \end{minipage}
    \hfill
    \begin{minipage}[t]{.245\textwidth}
        \centering
        \includegraphics[width=\textwidth]{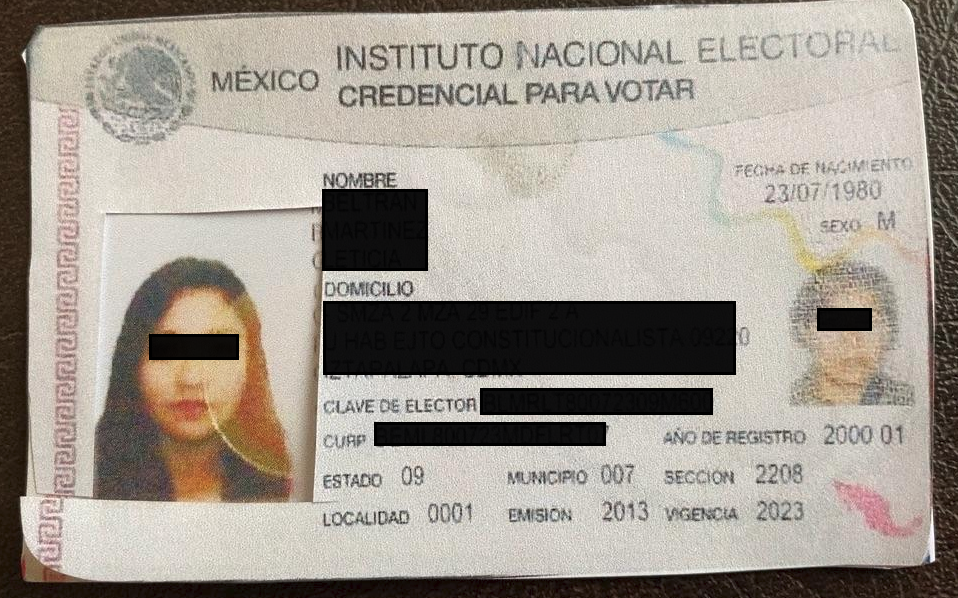}
        \subcaption{Mexican ID Card Composite.}\label{fig:10}
    \end{minipage}
    \hfill
    \begin{minipage}[t]{.24\textwidth}
        \centering
        \includegraphics[width=\textwidth]{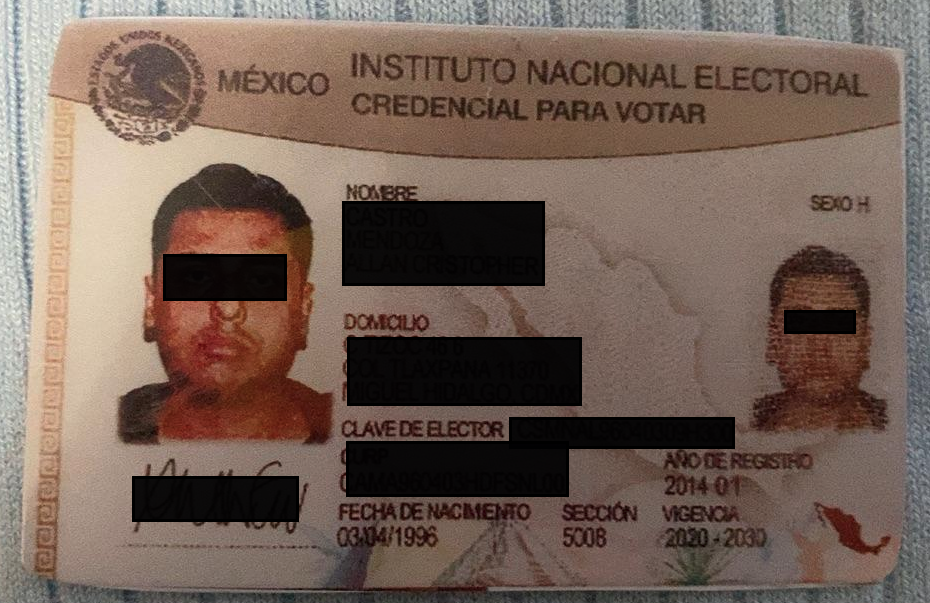}
        \subcaption{Mexican ID Card Printed.}\label{fig:11}
    \end{minipage}
    \hfill
    \begin{minipage}[t]{.24\textwidth}
        \centering
        \includegraphics[width=\textwidth]{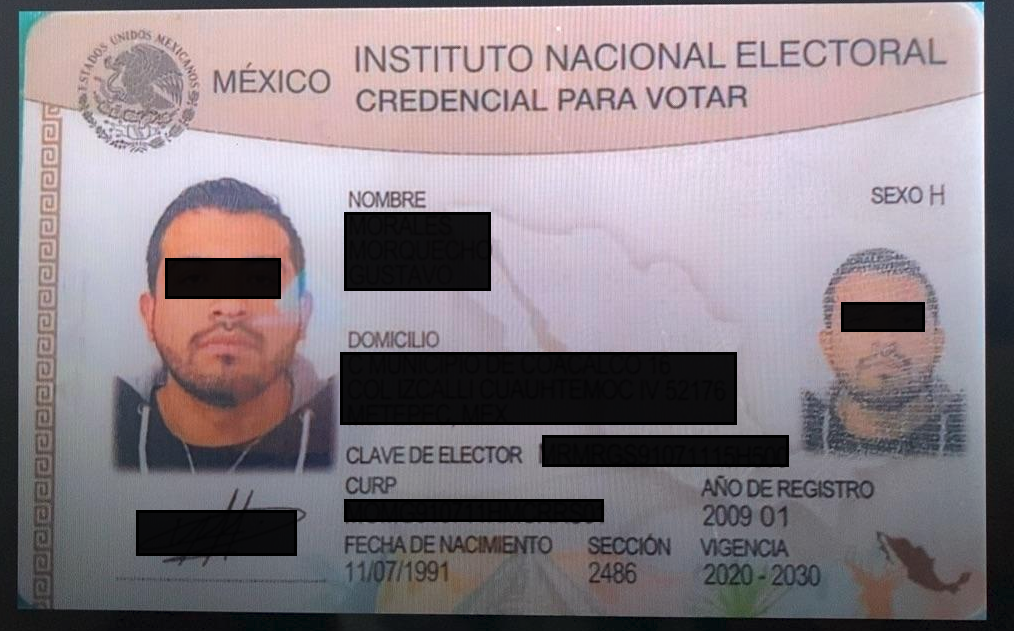}
        \subcaption{Mexican ID Card Display.}\label{fig:12}
    \end{minipage}
    \caption{Chilean, Argentinean and Mexican ID Cards examples. A tag was added in order to protect the person identity.}
\label{fig:1-2-3}
\end{figure*}

The size of the partitions for training, validation and test is shown in Table \ref{tab:DATASET}. All RGB images have a size of $1,280 \times 720$ pixels, and were acquired using different smartphones in real operation, with different illumination conditions, rotations, etc. 
The dataset contains four classes, according to their capture source: A digital image (Real), and three tampering classes: high quality printing (Printed on glossy paper), composed of segments from different ID Cards (Composite), and visualised on a screen (Display). All the images were captured in real face verification system with smartphones. Examples of the different classes for the Chile, Argentina and Mexico Cards are shown in Figure \ref{fig:1-2-3}.

\begin{table}[]
\label{tab:dataset}
\centering
\caption{Dataset Description}
\label{tab:DATASET}
\resizebox{0,48\textwidth}{!}{%
\begin{tabular}{|c|c|c|c|c|c|}
\hline
\textbf{ID Card}     & \textbf{Class} & \textbf{N IMG}  & \textbf{TRAIN}  & \textbf{VAL}   & \textbf{TEST} \\ \hline
                     & Composite      &  1,611          & 1,127            & 161            & 323          \\
\textbf{ARG1}        & Digital        & 2,129           & 1,490            & 212            & 427           \\
\textbf{}            & Printed        & 1,905           & 1,333            & 190            & 382          \\
\textbf{}            & Display        & 1,849           & 1,294            & 184            & 371          \\ 
\textbf{}            & Composite      & 2,335           & 1,634            & 233            & 468            \\
\textbf{ARG2}        & Digital        & 2,263           & 1,584            & 226            & 453            \\
                     & Printed        & 2,624           & 1,836            & 262            & 526            \\
                     & Display        & 2,177           & 1,523            & 217            & 437            \\ \hline
\textbf{}            & Composite      & 3,106           & 2,173           & 310             & 623             \\
\textbf{CHL1}        & Digital        & 3,444           & 2,410             & 344            & 690            \\
                     & Printed        & 3,027           & 2,118             & 302            & 607            \\
                     & Display        & 3,283           & 2,297             & 327            & 659            \\ 
                     & Composite      & 3,015           & 2,110            & 300            & 605            \\
\textbf{CHL2}        & Digital        & 2,986           & 2,087            & 298            & 601            \\
                     & Printed        & 3,003            & 2,101           & 299             & 603           \\
                     & Display        & 2,997           & 2,098            & 299             & 600           \\\hline 
                     & Composite      & 749             & 524             & 74            & 151             \\
\textbf{MEX}         & Digital         & 1,023              & 802            & 79            & 142             \\
                     & Printed        & 713             & 499             & 71             & 143            \\
                     & Display        & 768             & 537             & 76             & 155             \\\hline
                     & \textbf{Total} & \textbf{45,007} & \textbf{31,577} & \textbf{4,464} & \textbf{8,966} \\ \hline 
\end{tabular}%
}
\end{table}

\subsection{Manual Annotation}
\begin{figure}[H]
\centering
\includegraphics[scale=0.17]{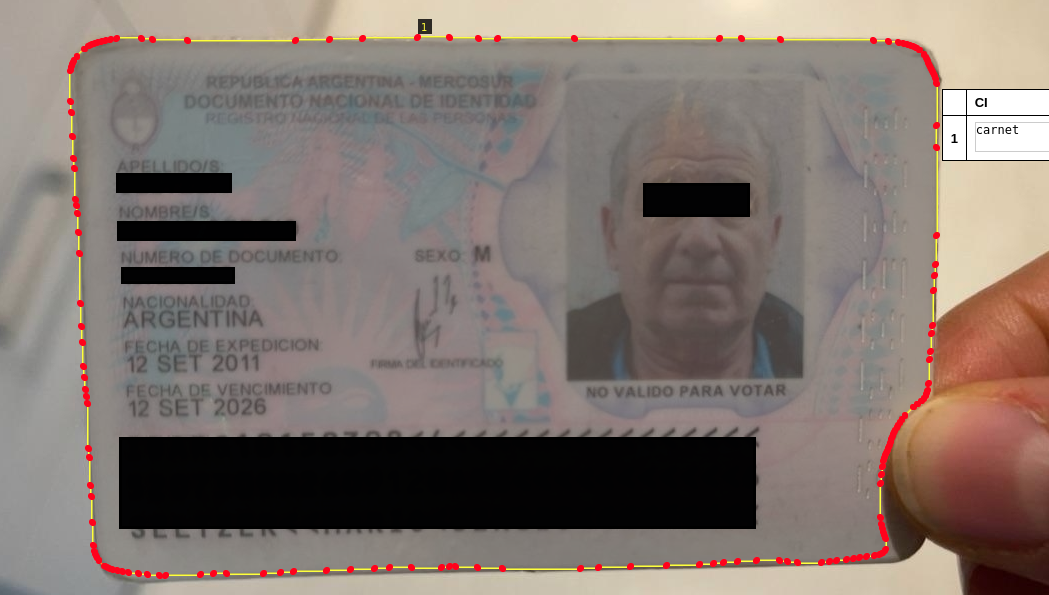}
\caption{ Example of manual annotation using VGG Image Annotator(VIA) 2.0.5.}
\label{fig_VIA}
\end{figure}

For training, we first manually annotate the boundaries of the ID Cards in the images, using the VIA Image Annotator version 2.0.5 \cite{VIA}. This is a very demanding task that allow us to estimate an IoU with high precision. An example of manual annotation using VIA is shown in Figure \ref{fig_VIA}.

\subsection{Preprocessing}
Three preprocessing methods were applied to the dataset, aimed to better represent the variability present in a real scenario, and obtain a model that generalizes the problem. Examples of preprocessing operations are shown in Figure \ref{dataug}.

\subsubsection{Background permuter}
\begin{figure}[H]
\centering
\includegraphics[scale=0.11]{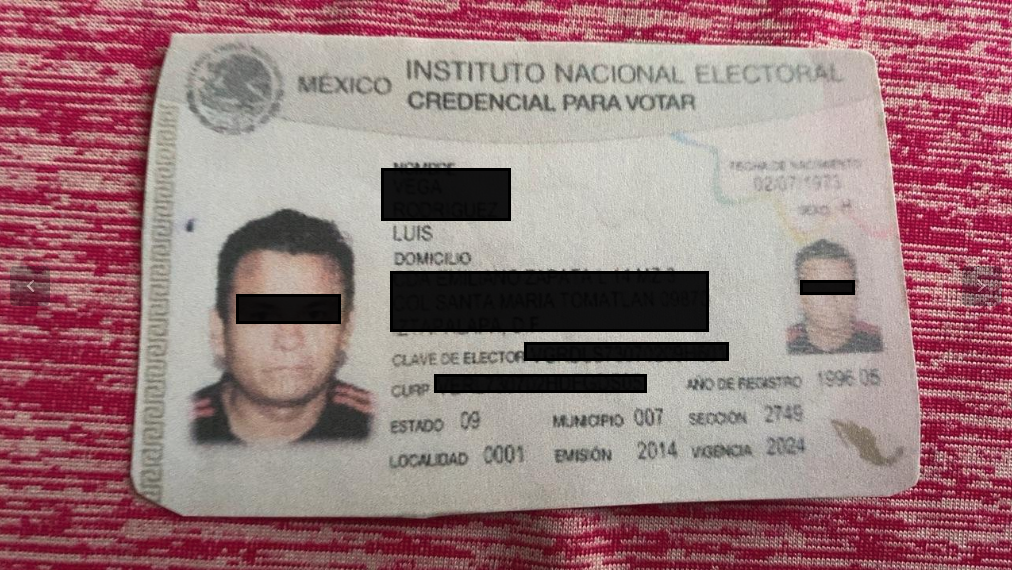}
\includegraphics[scale=0.10]{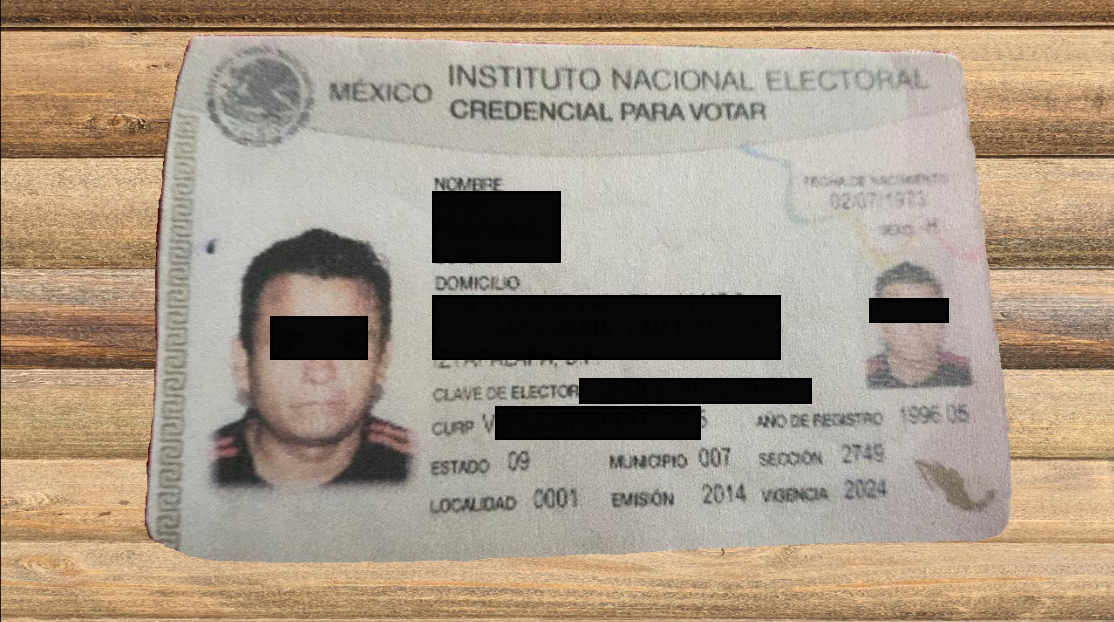}\\
\includegraphics[scale=0.10]{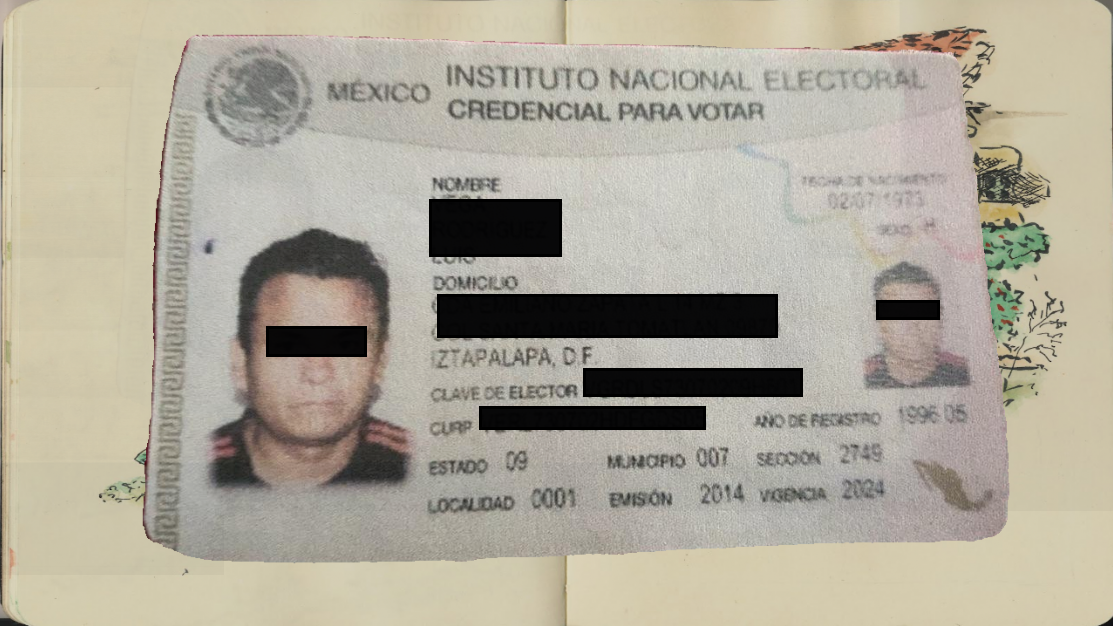}
\includegraphics[scale=0.10]{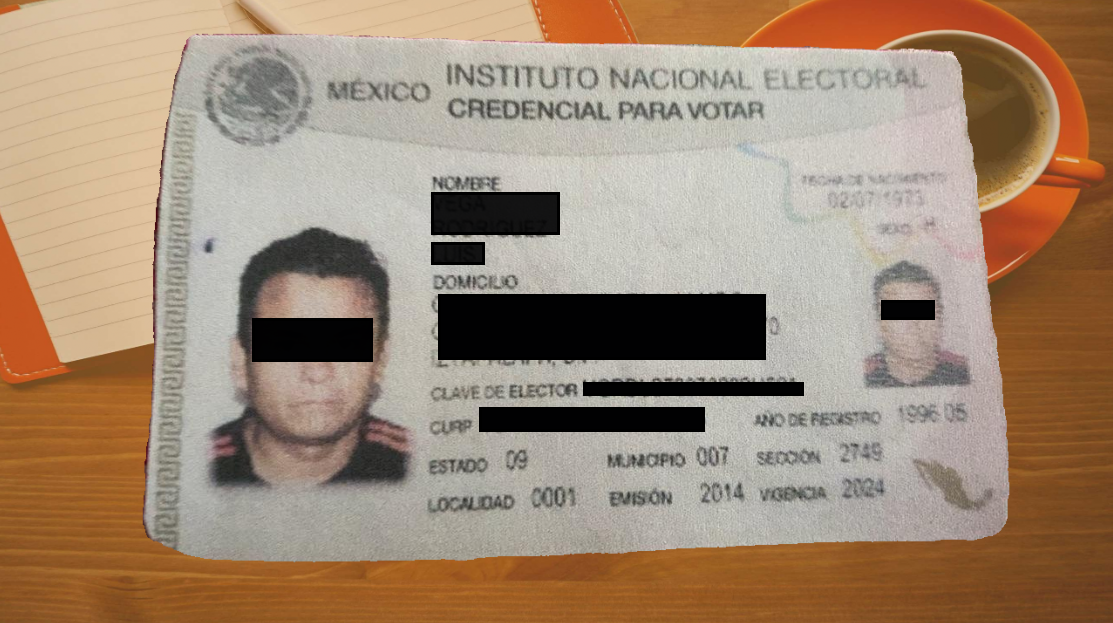}
\caption{Background permuter function: It can be observed that images with different backgrounds were created for the same ID Card.}
\label{fig:back_permuter}
\end{figure}

A lot of variation is present in real operation. In order to get a better representation of the problem, and to create challenging scenarios for training, a background function permuter was created. In this method, using the manually annotated mask for an ID Card, we select all pixels belonging the ID Card, and the we change the background, using different manually selected backgrounds, generating a new image, as shown in Figure \ref{fig:back_permuter}.

\subsubsection{Gray mask}
\begin{figure}[H]
\centering
\includegraphics[scale=0.13]{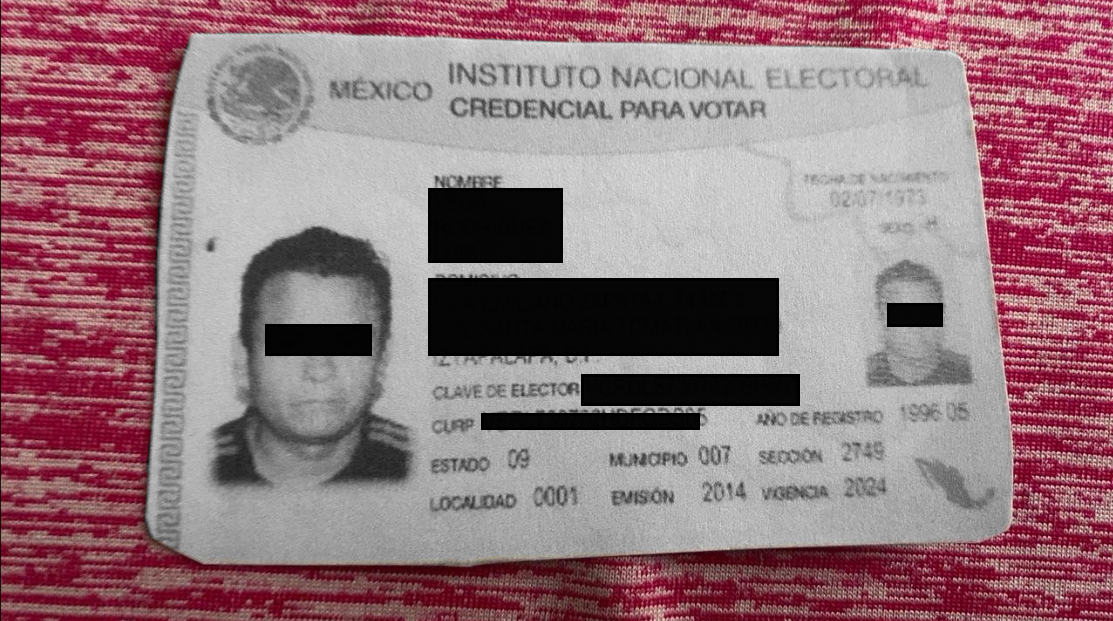}
\caption{ Gray mask: It can be observed that only the ID Card is in grayscale, while the background is in RGB. }
\label{gray_mask}
\end{figure}

This method uses the manually annotated mask for an ID Card, selecting all the pixels belonging to a card, and converting them from RGB to Gray, but leaving the background without any modification. Figure \ref{gray_mask} show an example.

\subsubsection{RGB2HSV}
\begin{figure}[H]
\centering
\includegraphics[scale=0.135]{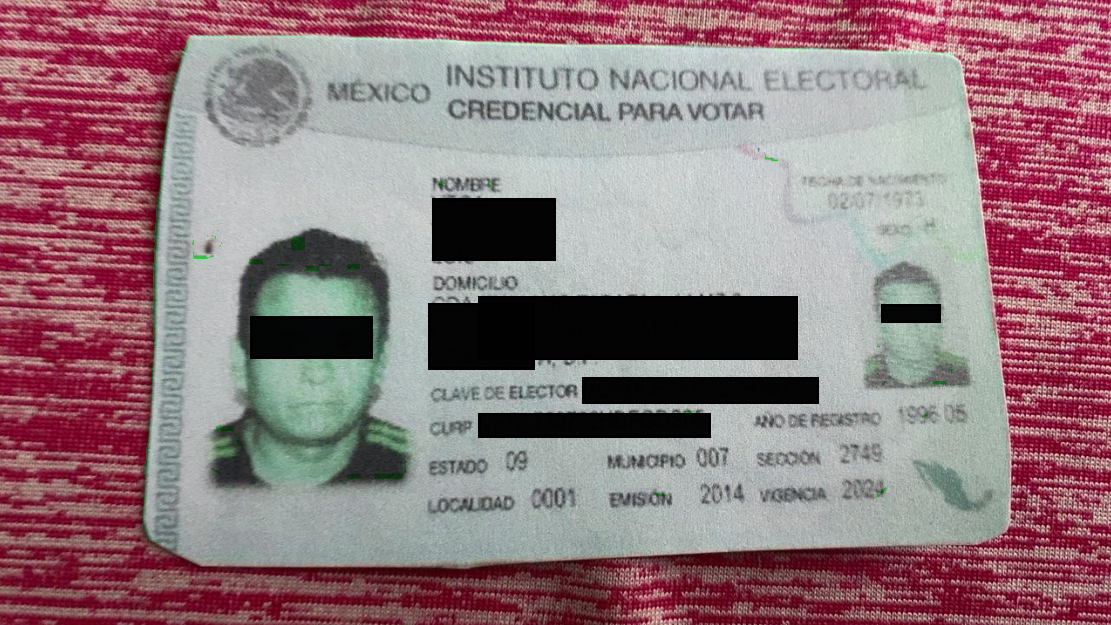}
\caption{ RGB to HSV: A variation in colour is performed changing the H and S channels in the HSV the colour space.}
\label{rgb2hsv}
\end{figure}

Like the two previous methods, here the manually annotated mask for an ID Card is used, selecting all the pixels belonging to a card, then a change in colour space is performed, from RGB to HSV, and then a random angle $H$ is chosen from the interval [-10º, 18º], and a random multiplier is chosen between $0.9$ and $1.18$ for the $S$ channel, then we performed the conversion back from HSV to RGB, obtaining a new image with a modified colour tonality. An example can be seen in Figure \ref{rgb2hsv}. 

\subsection{Data augmentation}
\begin{figure}[H]
\centering
\includegraphics[scale=0.38]{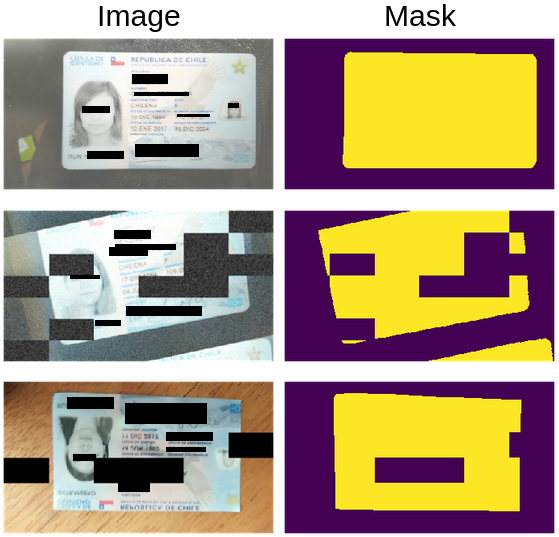}
\caption{Example of data augmentation applied.}
\label{dataug}
\end{figure}

Extensive data augmentation was performed on the dataset, in this way, we obtained many more images for training compared to the original size of the dataset. The operations performed and their respective parameters, using the implementations in the ImgAug library \cite{imgaug}. All the parameters used are reported in Table \ref{tab:DA-Param}. Examples are presented in Figure \ref{dataug}.

\begin{table}[H]
\centering
\caption{Functions and parameters of Data-Augmentation used for train models.}
\label{tab:DA-Param}
\resizebox{0.45\textwidth}{!}{%
\begin{tabular}{|c|c|}
\hline
\textbf{Function}                                                 & \textbf{Parameters}                                                                                             \\ \hline
Additive-Gaussian-Noise                                           & \textit{\begin{tabular}[c]{@{}c@{}}(loc = 0,  scale(0.0,  0.05  x  255),\\  per-channel = 0.5)\end{tabular}}    \\ \hline
Additive-Laplace-Noise                                            & \textit{(0.05  x  255)}                                                                                         \\ \hline
\begin{tabular}[c]{@{}c@{}}Additive-\\ Poisson-Noise\end{tabular} & \textit{(16.0)}                                                                                                 \\ \hline
Motion-Blur                                                       & \textit{(k = 3)}                                                                                                \\ \hline
\begin{tabular}[c]{@{}c@{}}AddToHueAnd\\Saturation\end{tabular} & \textit{(-50,  50)}                                                                                             \\ \hline
BilateralBlur                                                     & \textit{\begin{tabular}[c]{@{}c@{}}(d =  (3,  10),  sigma-color(10,\\ 250), sigma space(10, 250))\end{tabular}} \\ \hline
Coarse-Dropout                                                    & \textit{(p = (0.1, 0.35))}                                                                                      \\ \hline
Dropout2d                                                         & \textit{(p = 0.5)}                                                                                              \\ \hline
Edge-Detect                                                       & \textit{(alpha = (0.0, 0.7))}                                                                                   \\ \hline
\begin{tabular}[c]{@{}c@{}}Elastic-\\ Transformation\end{tabular} & \textit{(alpha = (0,   7.0), sigma = 0.25)}                                                                     \\ \hline
Gaussian-Blur                                                     & \textit{( sigma = 0.5)}                                                                                         \\ \hline
Spatter                                                           & \textit{(severity = 3)}                                                                                         \\ \hline
Rot180                                                             & \textit{([1,3])}                                                                                                \\ \hline
Flipud                                                            & \textit{(1)}                                                                                                  \\ \hline
Fliplr                                                            & \textit{(1)}                                                                                                  \\ \hline
\end{tabular}%
}
\end{table}

\subsection{Evaluation Metric - Intersection Over Union}
The performance metric for the segmentation accuracy performed by the trained models, was the Intersection Over Union (IoU), defined as follows for two arbitrary shapes $A$ and $B$:
\[ IoU = \frac{|A \cap B|}{|A \cup B|} \]

\begin{figure}[H]
\centering
\includegraphics[scale=0.25]{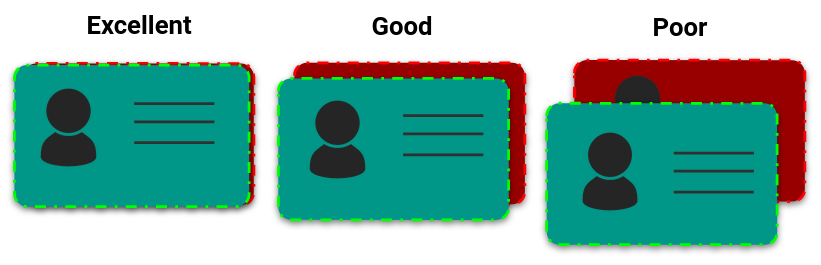}
\caption{ Intersection Over Union Example applied to ID Card Segmentation. }
\label{fig_IOU_metric}
\end{figure}

In our case, $A$ is the annotated mask, and $B$ is the predicted mask. This metric is widely used in the state-of-the-art. In Figure \ref{fig_IOU_metric}, examples for the IoU metric with ID Cards is shown.  

\section{Experiments and results}
\label{sec:Experiment}
In this section, segmentation performance obtained from models MobileUNet and DenseNet10 is compared using the IoU metric, and also the baseline model HOG/SVM is evaluated. Experiments were performed using two input resolutions, $224 \times 224$ px and $448 \times 448$ px.
 
\subsection{HOG/SVM}
A sliding window HOG/SVM object detector was used as a coarse segmentation, using its output as a baseline for comparison with more advanced methods, i.e., Deep Learning based methods. This decision is because the HOG/SVM method is currently being used as a basic ID Card segmentation in our current operating system. We needed a better performance approach because of the many errors that this method presented. 

\subsection{MobileUNet}
The MobileUNet model was trained using Tensorflow 2 \cite{tensorflow} as a framework, with Adam as optimizer \cite{adam} with a learning rate of 0.0001, and Categorical Crossentropy as loss function. Every training consist of 300 epochs and a batch size of 10. Extensive data augmentation was used.
 
\subsection{DenseNet}
This model was trained using Tensorflow 1.14, with Adam as optimizer with a learning rate of 0.0001, and Categorical Crossentropy as loss function. Every training consists of 300 epochs and a batch size of 10. The best results was obtained with grown rate $K=5$. Extensive data augmentation was also used.
 
\subsection{Results}
\label{sec:results}
The models were trained using the corresponding training partitions, and were evaluated using the test partitions corresponding to each country, containing 4,988 unique images for Chilean ID Cards, 3,387 unique images for Argentinean ID Cards and 591 unique images for Mexican ID Cards. A summary of the evaluation results for each method is shown in Table \ref{tb:summary_all}, where the best result was obtained on CHL subset, with a mean IoU of 0.9926, using MobileUNet with an input resolution of $448 \times 448$ px. Results for each country subset are shown in the following sections.

\subsubsection{Results-Chile}
\begin{figure}[H]
\centering
\includegraphics[scale= 0.23]{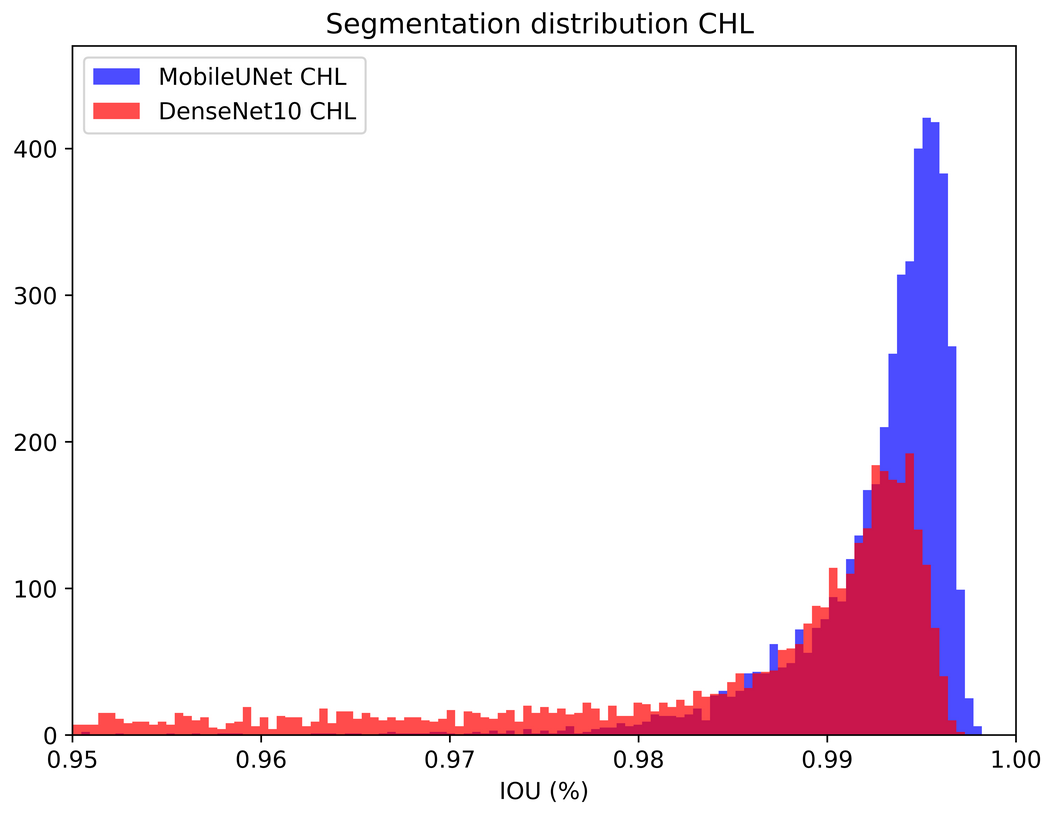}
\caption{Distribution Histogram for Chilean ID Card. }
\label{fig_Distribution_chl}
\end{figure}

Figure \ref{fig_Distribution_chl} shows the distribution of IoU scores for Chilean ID Card. This include the two kind of ID Cards available today in Chile (CHL1 and CHL2). An IoU of 0.9926 was obtained evaluating the MobileUNet model on the sum of both subsets.
Table \ref{tab:ChileID} shows the IoU distribution results for Chilean ID Cards, using a resolution of $448 \times 448$ px.

\begin{table}[H]
\centering
\scriptsize
\caption{Chilean ID Card Segmentation results. 75p represents percentile 75\%.}
\label{tab:ChileID}
\begin{tabular}{|c|c|c|c|c|c|c|}
\hline
\textbf{Method} & \textbf{Train}    & \textbf{Test} &  \textbf{Imgs test} & \textbf{mIoU}   & \textbf{Stdv}   & \textbf{75p}       \\ \hline
\textit{MobileUNet} & CHL & CHL1 &    2,579 &  0.9924    & 0.0052 & 0.9955     \\ 
    & & CHL2 &    2,409 &  0.9927    & 0.0056 & 0.9954     \\ \hline
\textit{DenseNet10} & CHL & CHL1 &    2,579 &  0.9875    & 0.0238 & 0.9938     \\ 
   & & CHL2 &    2,409 &  0.9883    & 0.0176 & 0.9939     \\ \hline
\end{tabular}
\end{table}

\subsubsection{Results-Argentina}
\begin{figure}[H]
\centering
\includegraphics[scale= 0.22]{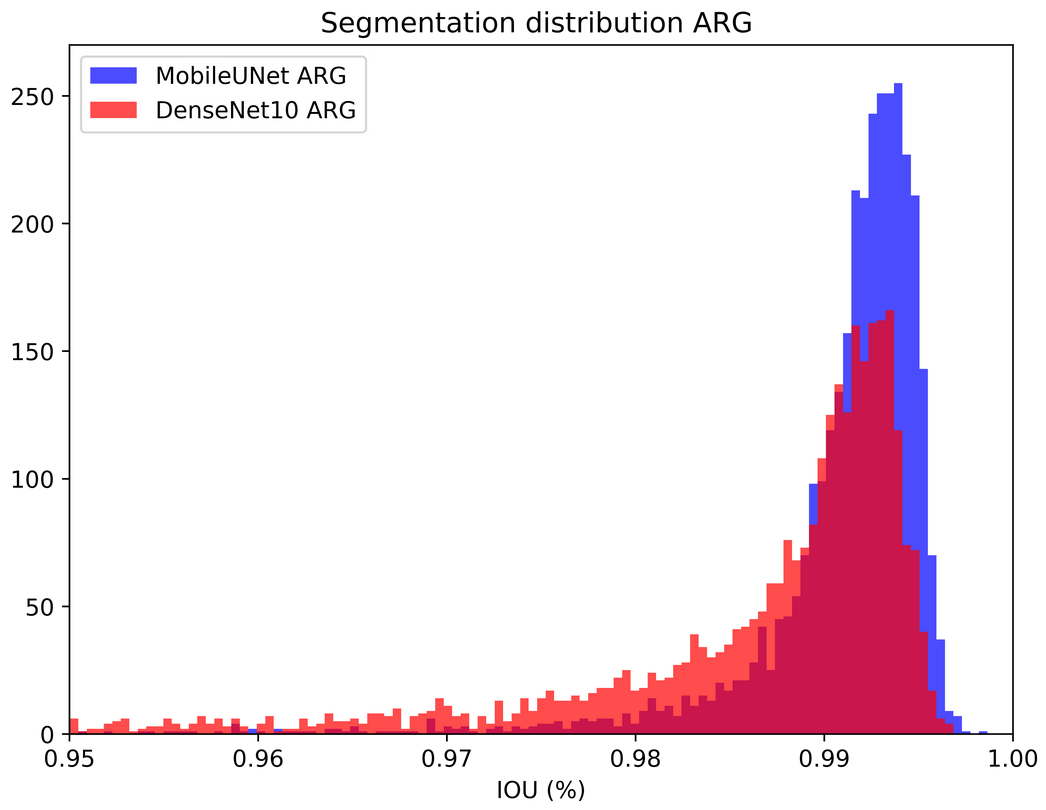}
\caption{Distribution Histogram for Argentinean ID Card.}
\label{fig_Distribution_arg}
\end{figure}

Figure \ref{fig_Distribution_arg} shows the distribution of IoU scores for Argentina ID Cards. This include the two kind of ID Card available today in Argentina (ARG1 and ARG2). An IoU of 0.9891 was obtained evaluating the MobileUNet model on the sum of both subsets. Table \ref{tab:Arg} shows the IoU distribution results for Argentina ID Cards, using a resolution of $448 \times 448$ px.
\vspace{-0.5cm}

\begin{table}[H]
\centering
\scriptsize
\caption{Argentinean ID Card Segmentation results.75p represents percentile 75\%.}
\label{tab:Arg}
\begin{tabular}{|c|c|c|c|c|c|c|}
\hline
\textbf{Method} & \textbf{Train}   & \textbf{Test}  & \textbf{Imgs test} & \textbf{mIoU}  & \textbf{Stdv}  & \textbf{75p}   \\ \hline
\textit{MobileUNet} & ARG & ARG1   &   1,503  & 0.9893 & 0.031 & 0.9941  \\ 
 & & ARG2   &   1,884  & 0.9889 & 0.010  & 0.9930 \\ \hline
\textit{DenseNet10} & ARG & ARG1   &   1,503  & 0.9709 & 0.055 & 0.9920  \\ 
  & & ARG2   &   1,884  & 0.9762 & 0.044 & 0.9925  \\ \hline
\end{tabular}
\end{table}

\subsubsection{Results-Mexico}

\begin{figure}[]
\centering
\includegraphics[scale= 0.21]{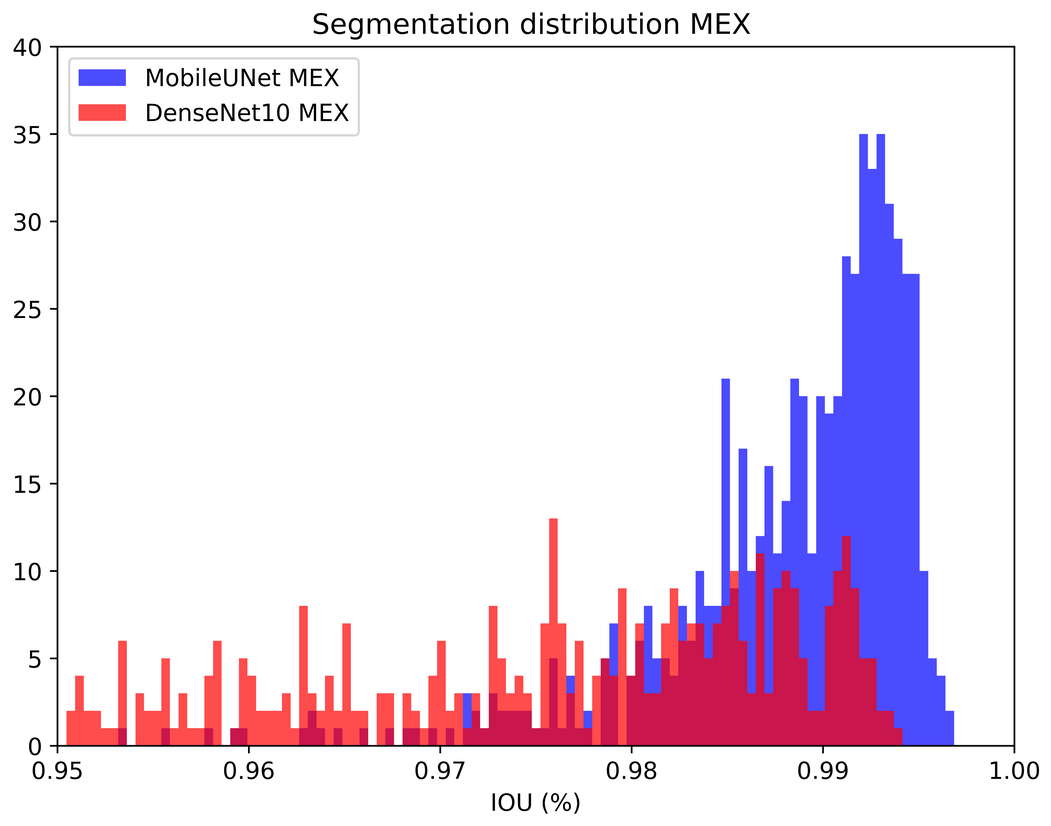}
\caption{Distribution Histogram for Mexican ID Card.}
\label{fig_Distribution_mex}
\end{figure}

Figure \ref{fig_Distribution_mex} shows the distribution of IoU scores for Mexico ID Cards. An IoU of 0.9862 was obtained evaluating the MobileUNet model. Table \ref{tab:mex} shows the IoU distribution results for Mexican ID Cards, using a resolution of $448 \times 448$ px.

\begin{table}[H]
\scriptsize
\centering
\caption{Mexican ID Card Segmentation results. 75p represents percentile 75\%.}
\label{tab:mex}
\begin{tabular}{|c|c|c|c|c|c|c|}
\hline
\textbf{Method}  & \textbf{Train}  & \textbf{Test}  & \textbf{Imgs test} & \textbf{mIoU} & \textbf{Stdv}   & \textbf{75p}       \\ \hline
\textit{MobileUNet} & MEX & MEX1 &   591      & 0.986& 0.0274 & 0.9928   \\ \hline
\textit{DenseNet10} & MEX & MEX1 &   591      & 0.9311 & 0.0928 & 0.9823  \\ \hline
\end{tabular}
\end{table}

\subsubsection{Results-ALL}
\begin{figure}[]
\centering
\includegraphics[scale= 0.23]{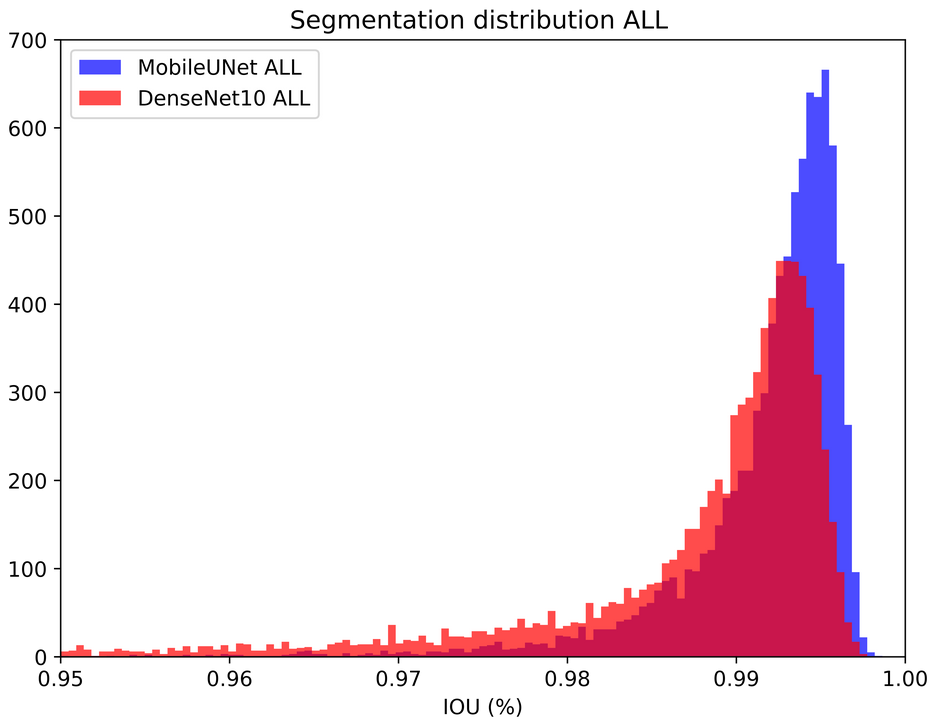}
\caption{Distribution Histogram for fusion multi-country data of Chile, Argentina and Mexico. }
\label{fig:summary_all}
\end{figure}

Figure \ref{fig:summary_all} shows the distribution of IoU scores using a multi-country dataset of Chile, Argentina and Mexico ID Cards for training. This include the five kind of ID Card available today in the three countries. An IoU of 0.9911 was obtained evaluating the MobileUNet model on all the subsets. Table \ref{tab:All} shows the IoU distribution results for all the ID Cards, using a resolution of $448 \times 448$ px.

\begin{table}[H]
\centering
\scriptsize
\caption{Argentinean, Chilean and Mexican ID Cards segmentation results. 75p represents percentile 75\%.}
\label{tab:All}
\begin{tabular}{|c|c|c|c|c|c|c|}
\hline
\textbf{Method} & \textbf{Train}    & \textbf{Test}  & \textbf{Imgs test} & \textbf{mIoU}  & \textbf{Stdv}  & \textbf{75p}   \\ \hline
          & & ARG1   &   1,503  & 0.9908 & 0.031 & 0.9955  \\ 
          & & ARG2   &   1,884  & 0.9898 & 0.010  & 0.9935 \\ 
\textit{MobileUNet}& ALL & CHL1   &   2,579  & 0.9922 & 0.007 & 0.9952  \\ 
         &  & CHL2   &  2,409  & 0.9921 & 0.005 & 0.9954  \\ 
         &  & MEX1   &   591  & 0.9878 & 0.013 & 0.9968  \\ \hline
         &   & ARG1   &   1,503  & 0.9779 & 0.040 & 0.9923  \\ 
         &  & ARG2   &   1,884  & 0.9822 & 0.028  & 0.9928 \\ 
\textit{DenseNet10} & ALL & CHL1   &   2,579  & 0.9823 & 0.041 & 0.9935  \\ 
        &   & CHL2   &  2,409  & 0.9840 & 0.028 & 0.9939  \\ 
        &  & MEX1   &   591  & 0.9750 & 0.040 & 0.9914  \\ \hline
\end{tabular}
\end{table}

\begin{table*}[]
\centering
\scriptsize
\caption{Inference time of Semantic segmentation results for Chilean, Argentina and Mexican ID Cards.}
\label{tb:summary_all}
\begin{tabular}{|c|c|c|c|c|c|c|c|c|}
\hline
\textbf{Method}    &  \textbf{Input shape} & \textbf{Params} & \textbf{Country} & \textbf{mean IoU} & \textbf{std} & \textbf{Inf.Time} \\ \hline
\textit{SVM/HOG}  & 224 x 224  & N/A   & CHL & 0.8671   & 0.0705 & 0,004 s                 \\
                  & 448 x 448  & N/A   & CHL & 0.8913   & 0.0435 & 0.018 s                 \\ \hline
                  & 224 x 224  & 0,2 M & CHL & 0.9832   & 0.0113 & 0.026 s      \\
                  & 448 x 448  & 0,2 M & CHL & 0.9879   & 0.0237 & 0.034 s                 \\ 
\textit{DenseNet10}& 448 x 448  & 0,2 M & ARG & 0.9739   & 0.0492 & 0.033 s                 \\ 
                  & 448 x 448  & 0,2 M & MEX & 0.9311   & 0.0274 & 0.035 s                 \\ 
                  & 448 x 448  & 0,2 M & ALL & 0.9814   & 0.035 & 0.034 s                 \\ \hline
                  & 224 x 224  & 6,5 M & CHL & 0.9898  & 0.0053  & 0.024 s                 \\
                  & \textbf{448 x 448}  & \textbf{6,5 M} & \textbf{CHL} & \textbf{0.9926}   & \textbf{0.0054} & \textbf{0.023} s                 \\
\textit{MobileUNet}
                  & 448 x 448  & 6,5 M & ARG & 0.9891   & 0.0223 & 0.027 s                 \\
                  
                  & 448 x 448  & 6,5 M & MEX & 0.9862   & 0.0275 & 0.026 s                 \\
                  
                  & 448 x 448  & 6,5 M & ALL & 0.9911   & 0.0150 & 0.023 s                 \\ \hline
\end{tabular}%
\end{table*}

The best results were marginally better in the Chilean part of the dataset, probably because the Chilean ID Cards are ICAO compliant, this implies that the layout is fixed in the entire partition, the printing quality is better, the card is completely rigid, made from plastic, obtaining overall better quality images captured. We also can observe that the results on Argentina and Mexico present a high mean IoU, showing the robustness of this method in front of more difficult, non ICAO compliant ID Card types. Table \ref{tb:summary_all} shows the summary of all methods including the inference time. The inference time was estimated using 100 random images. For Deep Learning based methods, the lower inference time was obtained by MobileUNet with 0.023 seconds. It is important to notice that HOG/SVM performed notoriously worse in the Chilean subset, obtaining a difference of at least 10\% compared to the best Deep Learning approach, even if the inference time is lower.

\begin{figure}[H]
\centering
\includegraphics[scale= 0.16]{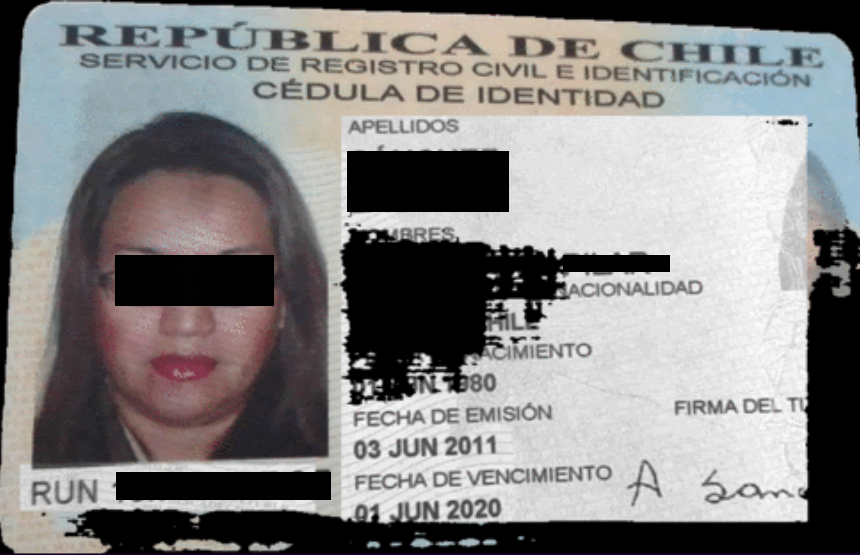}
\includegraphics[scale= 0.16]{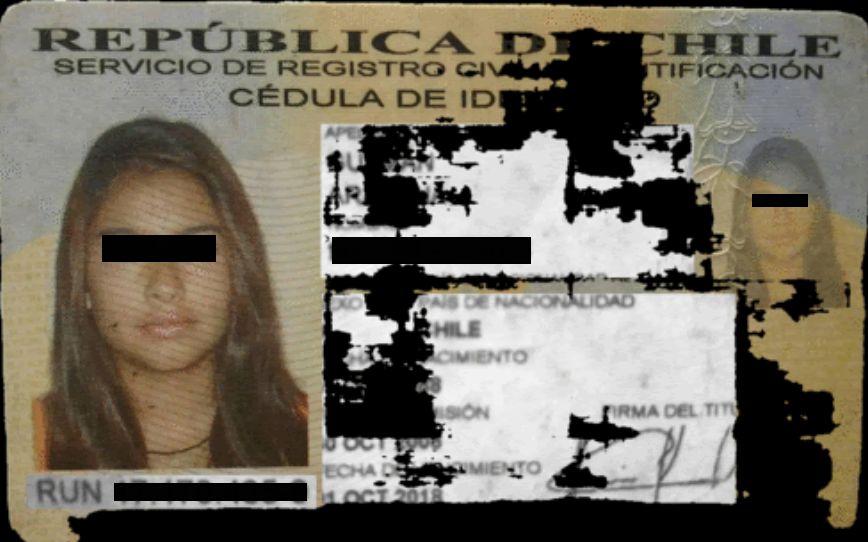}
\caption{Example of a low IoU composite scenario image with high segmentation error. The blue rectangle tags were added to protect the identity.}
\label{fig:border_error}
\end{figure}

Figure \ref{fig:border_error} shows an example of a Composite Chilean ID Card with a low IoU, presenting high segmentation error. This image shows a composite image with fake information on the ID Card. This scenario is one of the most basic attempts to try to fool the remote verification system. The artificial borders can confuse the segmentator. Black pixels represent background pixels (wrong detected).

\section{Conclusion}
\label{sec:conclusion}

In this work we investigated the performance of two semantic segmentation methods for ID Card images, with cluttered backgrounds and occlusions, so it can help to the following stages in a identity verification or document tampering detection system. The methods proposed were based on a MobileUNet and a lightweight version of DenseNet, where both methods showed good results on a private testing dataset, consisting of 8,966 images, that includes five different ID Card types from Chile, Argentina and Mexico, with real and presentation attack images. The best results were obtained using the MobileUNet model, with an input resolution of $448 \times 448$ px, yielding a mean IoU of $0.9911$ for the entire dataset. The best subset was Chile, obtaining a mean IoU of $0.9926$ using MobileUNet method, explained by the more consistent layout, because this subset is ICAO compliant, outperforming vastly the HOG/SVM approach used as a baseline. The results on Argentina and Mexico were marginally lower than Chile, but the images are qualitatively lower and the layout is not consistent, for example, the text fields and photo are in different positions in both Argentinean subsets. This implies that our method can be extended to other difficult types of ID Cards, without losing performance. As we mentioned before, MobileUNet reached better results, however this method has 6.5 millions of parameters, compared to DenseNet10, with only 210,732 parameters, and its results are still competitive. As a future work, we are working with others lightweight implementations such as EfficientNet or MobileNetV3, in order to reduce even more the number of parameters and the complexity of the models. This work can be used as a guide for future research efforts in this topic.

\section*{Acknowledgement}
This work was partially supported by TOC Biometrics, the German Federal Ministry of Education and Research, the Hessian Ministry of Higher Education, Research, Science and the Arts within their joint support of the National Research Center for Applied Cybersecurity ATHENE.

\bibliographystyle{ieeetr}
\bibliography{references}
\vspace{-1.0cm}

\begin{IEEEbiography}[{\includegraphics[width=1in,height=1.25in,clip,keepaspectratio]{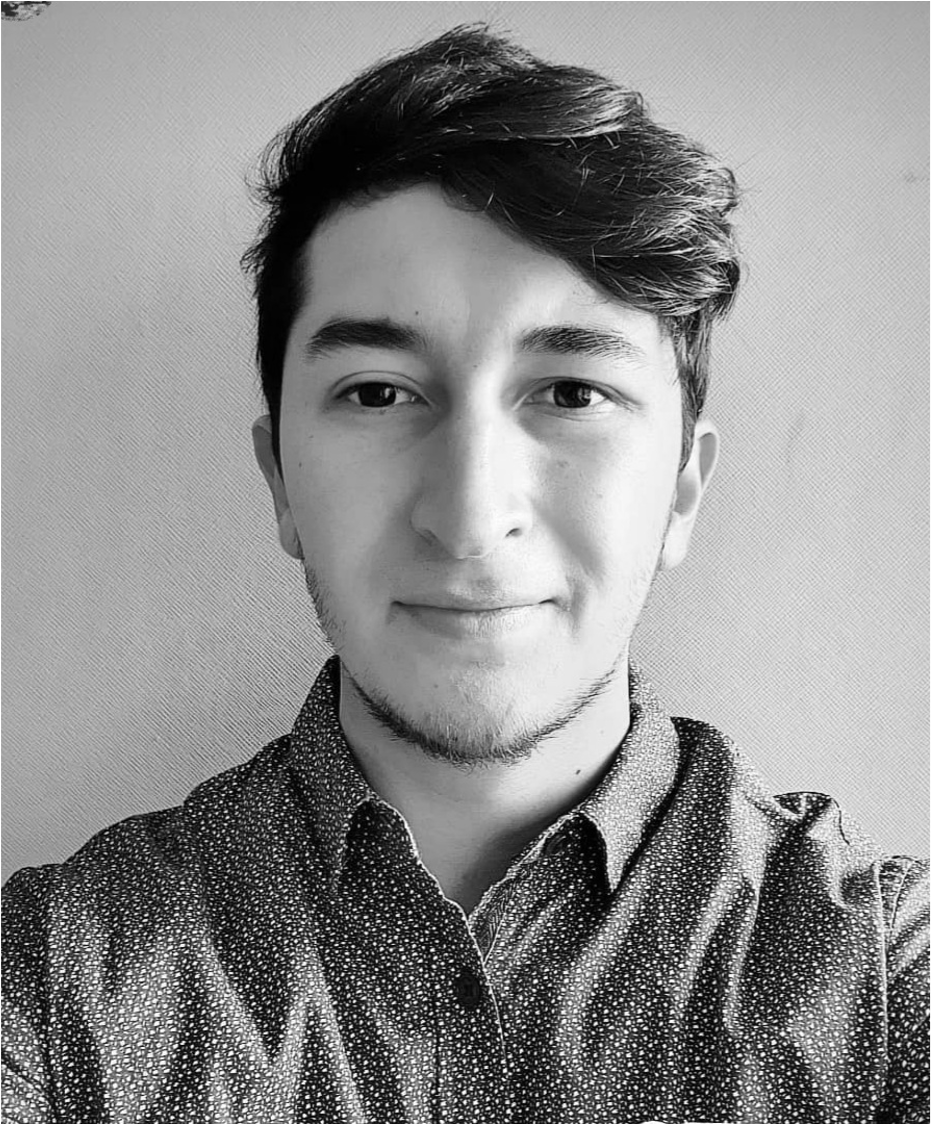}}]{Rodrigo Lara} received a B.S. in Computer Engineering from Universidad Andres Bello in 2020. Currently, he is a researcher at the R\&D center in TOC Biometrics. His main interests include computer vision, pattern recognition and deep learning applied to real problems such as tampering detection and semantic segmentation.
\end{IEEEbiography}

\begin{IEEEbiography}[{\includegraphics[width=1in,height=1.25in,clip,keepaspectratio]{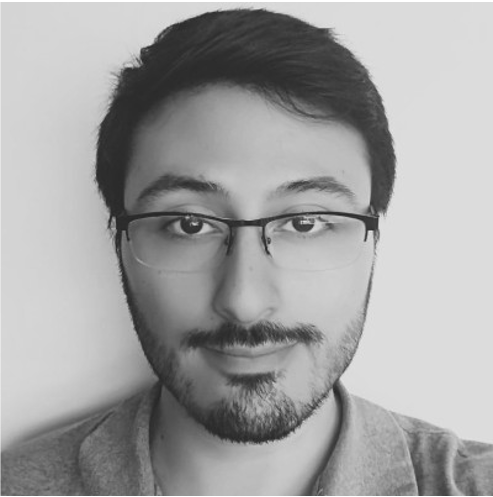}}]{Andres Valenzuela} received a B.S. in Computer Engineering from Universidad Andres Bello in 2019. Currently, he is researcher at the R\&D center in TOC Biometrics. His main interests include computer vision, pattern recognition and deep learning applied to real problems such as tampering detection and semantic segmentation.
\end{IEEEbiography}

\begin{IEEEbiography}[{\includegraphics[width=1in,height=1.25in,clip,keepaspectratio]{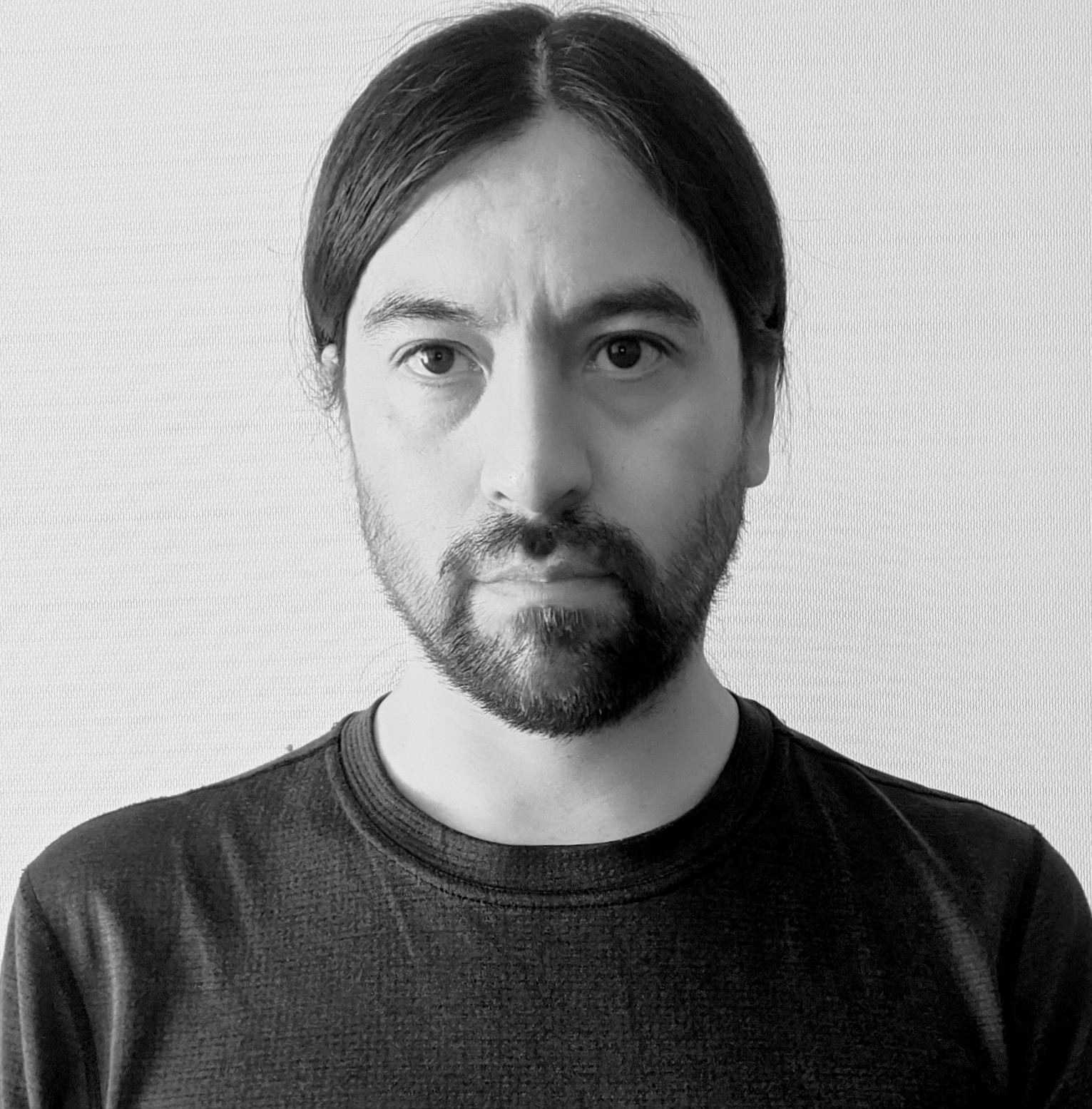}}]{Daniel Schulz} is a Ph.D. candidate from the Department of Electrical Engineering, Universidad de Chile, Santiago, Chile. He received the B.E. degree (Computer Science) from the Faculty of Engineering, Universidad Austral de Chile, in Valdivia, Chile, 2005. He is currently a Researcher at the R\&D center in TOC Biometrics. His main research interests are Biometrics, Computer Vision applied to Mining and Trademark Image Retrieval.
\end{IEEEbiography}

\begin{IEEEbiography}[{\includegraphics[width=1in,height=1.25in,clip,keepaspectratio]{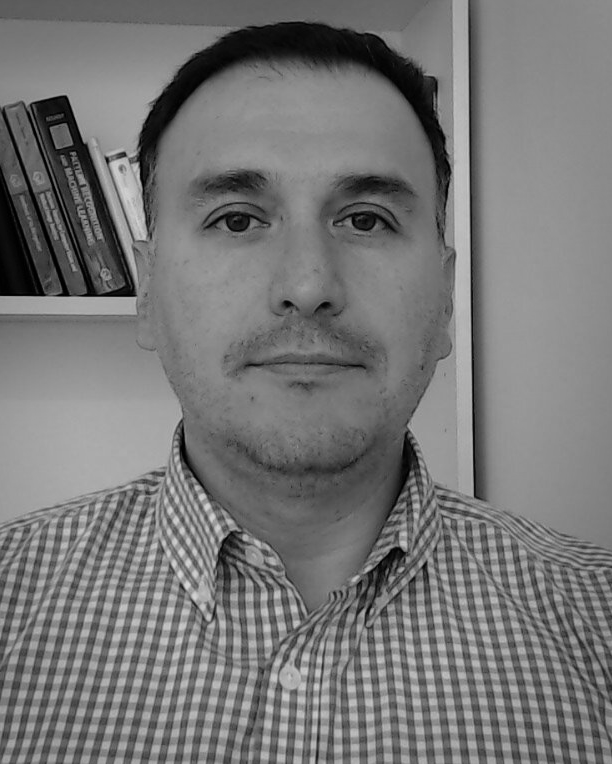}}]{Juan Tapia} received a P.E. degree in Electronics Engineering from Universidad Mayor in 2004, a M.S. in Electrical Engineering from Universidad de Chile in 2012, and a Ph.D. from the Department of Electrical Engineering, Universidad de Chile in 2016. In addition, he spent one year of internship at University of Notre Dame. In 2016, he received the award for best Ph.D. thesis. From 2016 to 2017, he was an Assistant Professor at Universidad Andres Bello. From 2018 to 2020, he was the R\&D Director for the area of Electricity and Electronics at Universidad Tecnologica de Chile. He is currently a Senior Researcher at Hochschule Darmstadt (HDA), and R\&D Director of TOC Biometrics. His main research interests include pattern recognition and deep learning applied to iris biometrics, morphing, feature fusion, and feature selection. 
\end{IEEEbiography}

\begin{IEEEbiography}[{\includegraphics[width=1in,height=1.25in,clip,keepaspectratio]{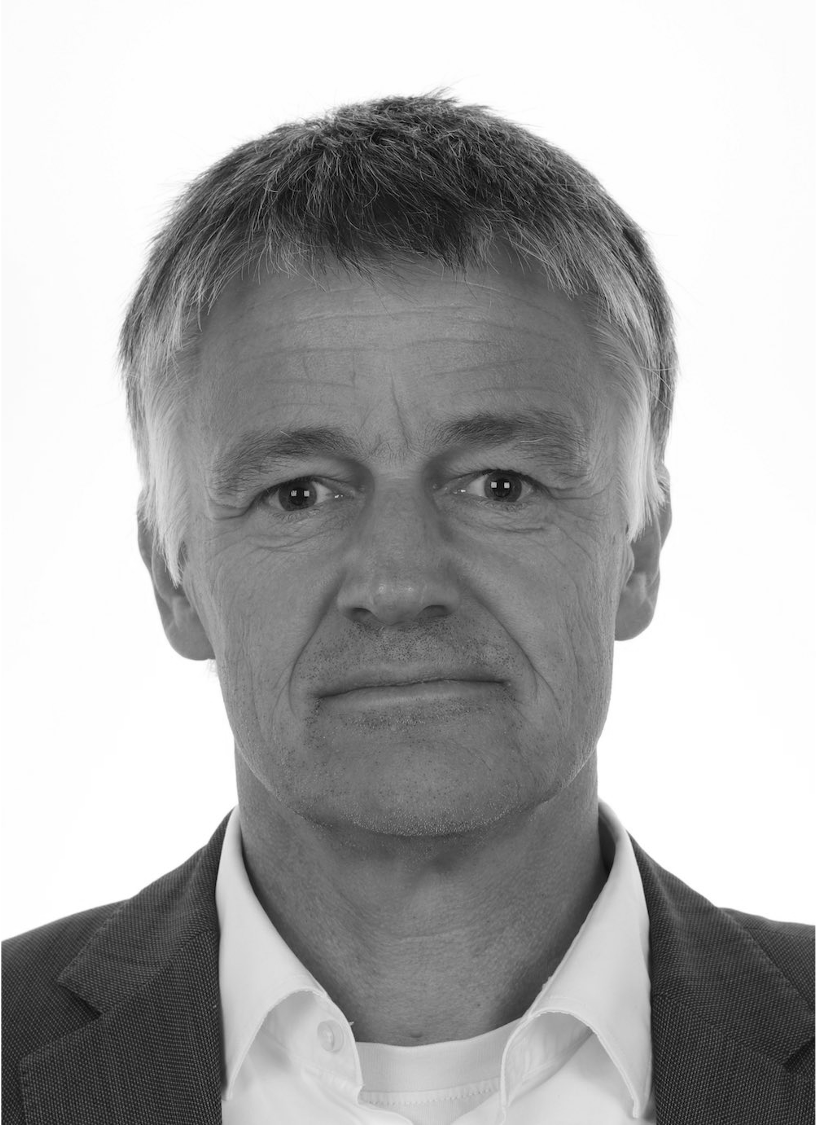}}]{Christoph Busch} is member of the Department of Information Security and Communication Technology (IIK) at the Norwegian University of Science and Technology (NTNU), Norway. He holds a joint appointment with the computer science faculty at Hochschule Darmstadt (HDA), Germany. Further he lectures the course Biometric Systems at Denmark’s DTU since 2007. On behalf of the German BSI he has been the coordinator for the project series BioIS, BioFace, BioFinger, BioKeyS Pilot-DB, KBEinweg and NFIQ2.0. In the European research program he was initiator of the Integrated Project 3D-Face, FIDELITY and iMARS. Further he was/is partner in the projects TURBINE, BEST Network, ORIGINS, INGRESS, PIDaaS, SOTAMD, RESPECT and TReSPAsS. He is also principal investigator in the German National Research Center for Applied Cybersecurity (ATHENE). Moreover Christoph Busch is co-founder and member of board of the European Association for Biometrics (www.eab.org) that was established in 2011 and assembles in the meantime more than 200 institutional members. Christoph co-authored more than 500 technical papers and has been a speaker at international conferences. He is member of the editorial board of the IET journal.
\end{IEEEbiography}
\end{document}